\documentclass[10pt,twocolumn,letterpaper]{article}
\usepackage{cvpr}
\usepackage{graphicx}
\usepackage{amsmath}
\usepackage{amssymb}
\usepackage{booktabs}
\usepackage{algpseudocode}
\usepackage{algorithm}
\usepackage{mathtools}
\usepackage{stmaryrd}
\usepackage{trimclip}
\usepackage[dvipsnames]{xcolor}

\makeatletter
\DeclareRobustCommand{\shortto}{%
  \mathrel{\mathpalette\short@to\relax}%
}

\newcommand{\short@to}[2]{%
  \mkern2mu
  \clipbox{{.5\width} 0 0 0}{$\m@th#1\vphantom{+}{\shortrightarrow}$}%
  }
\makeatother
\newcommand{\ShortName}{DyNFL\xspace}
\newcommand{\method}{DyNFL\xspace}

\newcommand{\pos}{\mathbf{p}}

\newcommand{\x}{\mathbf{x}}
\newcommand{\dir}{\mathbf{d}}
\newcommand{\origin}{\mathbf{o}}

\newcommand{\real}{\mathbb{R}}
\newcommand{\ray}{\mathbf{r}}
\newcommand{\density}{\sigma}
\newcommand{\opacity}{\alpha}
\newcommand{\reflectance}{\rho}

\newcommand{\intensity}{e}
\newcommand{\pdrop}{p_d}

\newcommand{\leik}{\mathcal{L}_{\text{eik}}}
\newcommand{\trans}{\mathcal{T}}
\newcommand{\bwr}{\includegraphics[width=3em,height=0.8em]{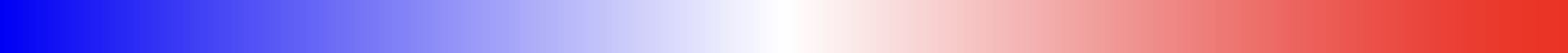}}
\newcommand{\exponential}[1]{\text{exp}\left(#1\right)}

\newcommand{\posfeat}{\mathbf{f}_{\text{pos}}}
\newcommand{\geofeat}{\mathbf{f}_{\text{geo}}}

\newcommand{\dirfeat}{\mathbf{f}_{\text{dir}}}

\newcommand{\rayfeat}{\mathbf{f}_{\text{ray}}}

\newcommand{\reflectivity}{\rho}
\newcommand{\weight}{\alpha}

\definecolor{cvprblue}{rgb}{0.21,0.49,0.74}
\usepackage[pagebackref,breaklinks,colorlinks,citecolor=cvprblue]{hyperref}

\usepackage[capitalize]{cleveref}
\crefname{section}{Sec.}{Secs.}
\Crefname{section}{Section}{Sections}
\Crefname{table}{Table}{Tables}
\crefname{table}{Tab.}{Tabs.}

\def\paperID{2088} 
\def\confName{CVPR}
\def\confYear{2024}

\title{Dynamic LiDAR Re-simulation using Compositional Neural Fields}

\author{
Hanfeng Wu$^{1,2}$ \quad Xingxing Zuo$^{2,}$\footnotemark[1]\quad Stefan Leutenegger$^{2}$\\ 
Or Litany$^{3,4}$\quad Konrad Schindler$^{1}$ \quad Shengyu Huang$^{1,}$\footnotemark[1]\\
{\small $^{1}$ ETH Zurich \quad $^{2}$ TU Munich \quad $^{3}$ Technion\quad $^{4}$ NVIDIA
}
}

\begin{document}
\renewcommand{\thefootnote}{*}
\maketitle
\footnotetext[1]{Corresponding authors}
\begin{figure*}[t]
    \centering
        \includegraphics[width=1.0\textwidth]{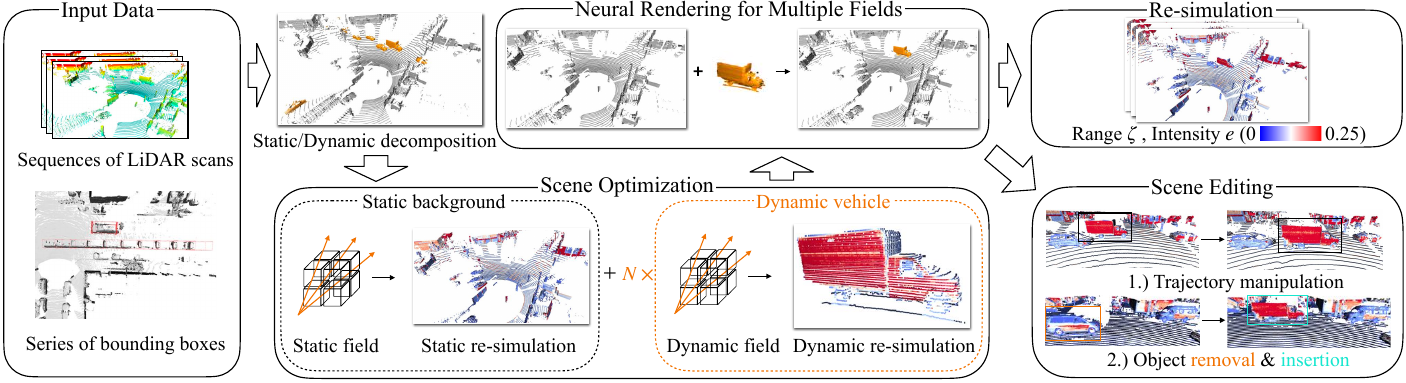}
        \vspace{-4mm}
        \caption{
        Overview of \method. Our method takes LiDAR scans and tracked bounding boxes of dynamic vehicles as input. \method first decomposes the scene into a static background and $N$ dynamic vehicles, each modelled using a dedicated neural field. These neural fields are then composed to re-simulate LiDAR scans in dynamic scenes. Our composition technique supports various scene edits, including altering object trajectories, removing and adding reconstructed neural assets between scenes.
    }
    \label{fig:main}
\end{figure*}
\begin{abstract}
We introduce~\ShortName, a novel neural field-based approach for high-fidelity re-simulation of LiDAR scans in dynamic driving scenes. \ShortName processes LiDAR measurements from dynamic environments, accompanied by bounding boxes of moving objects, to construct an editable neural field. This field, comprising separately reconstructed static background and dynamic objects, allows users to modify viewpoints, adjust object positions, and seamlessly add or remove objects in the re-simulated scene. 
A key innovation of our method is the neural field composition technique, which effectively integrates reconstructed neural assets from various scenes through a ray drop test, accounting for occlusions and transparent surfaces.
Our evaluation with both synthetic and real-world environments demonstrates that \ShortName substantially improves dynamic scene LiDAR simulation, offering a combination of physical fidelity and flexible editing capabilities. \href{https://shengyuh.github.io/dynfl}{[project page]}

\end{abstract}


\section{Introduction}
We introduce a neural representation for the purpose of reconstructing and manipulating LiDAR scans of dynamic driving scenes. 
Counterfactual re-simulation is an emerging application in the realm of autonomous driving, offering a unique approach to examining "what if" scenarios. This method involves creating a reconstruction of a real-world event, termed as \textit{digital twin}, and then applying various modifications to it. These could include altering the environmental conditions, changing the actions of some agents, or introducing additional scene elements. Analyzing the outcomes of these edited scenarios provides insights into the functioning of the perception system, moreover they can be used to obtain training data for rare situations.

The essence of counterfactual re-simulation is the capability to authentically recreate variations of the original, factual observation. We address this challenge in the context of LiDAR on autonomous vehicles (AV). Existing approaches to LiDAR re-simulation have important limitations. Conventional simulators such as CARLA~\cite{dosovitskiy2017carla} and NVIDIA DRIVE Sim are capable of modeling LiDAR sensors. However, their reliance on manually designed 3D simulation assets requires significant human effort. LiDARsim~\cite{manivasagam2020lidarsim} aims to remedy this by reconstructing vehicles and scenes from real measurements. While producing encouraging results, its two-stage LiDAR modeling lacks realism, particularly in terms of physical effects like multi-returns and reflected intensity, which were shown to matter for downstream processing~\cite{guillard2022learning}. Following NeRF's~\cite{mildenhall2020nerf} success in camera view synthesis, some works have applied neural fields for LiDAR modeling~\cite{Huang2023nfl, tao2023lidar, zhang2023nerf}. In particular, Neural LiDAR Fields (NFL)~\cite{Huang2023nfl} offer a physically inspired LiDAR volumetric rendering scheme that accounts for two-way transmittance and beam width, allowing faithful recovery of secondary returns, intensity, and ray drops. These models are, however, limited to static scenes that do not change while multiple input views are scanned, and are thus of limited use for re-simulation in the presence of moving traffic. Recently, UniSim~\cite{yang2023unisim} followed Neural Scene Graph~\cite{Ost_2021_CVPR} in modeling road scenes as sets of movable NeRF instances on top of a static background. UniSim introduced a unified synthesis approach for camera and LiDAR sensors, but ignored physical sensor properties like two-way transmittance and beam width~\cite{Huang2023nfl}.

We present \ShortName, a novel approach for re-simulating LiDAR views of driving scenarios. Our method builds upon a neural SDF that enables an accurate representation of scene geometry, while at the same time enforcing physical accuracy by modeling two-way transmittance, like NFL~\cite{Huang2023nfl}. 
Our primary contribution is a method for compositing neural fields that accurately integrates LiDAR measurements from individual fields representing different scene assets. With the help of a ray drop test, we effectively manage occlusions and transparent surfaces. This not only ensures physical accuracy, but also facilitates the inclusion of assets reconstructed from a variety of static and dynamic scenes, thereby enhancing control over the simulated content. Our method bridges the gap between the physical fidelity of the re-simulation and the flexible editing of dynamic scenes.
We validate \ShortName with both synthetic and real-world data, focusing on three key areas: \textit{(i)} high-quality view synthesis, \textit{(ii)} perceptual fidelity, and \textit{(iii)} asset manipulation. We find that our approach outperforms baseline models in terms of both range and intensity estimates. Its synthetic outputs also show higher agreement with real scans on object detection and segmentation tasks. \ShortName enables not only removal, duplication and repositioning of assets within the same scene, but also the inclusion of assets reconstructed in other scenes, paving the way for new applications.

\section{Related work}
\subsection{Neural radiance fields and volume rendering} 
Neural Radiance Fields (NeRF)~\cite{mildenhall2020nerf} have demonstrated remarkable success in novel-view image synthesis through neural volume rendering. These fields are characterized by the weights of Multilayer Perceptrons (MLPs), which enable the retrieval of volume density and RGB colors at any specified point within the field for image compositing via volume rendering. Several studies~\cite{barron2021mip,barron2022mip,verbin2022ref,chen2022tensorf,fridovich2022plenoxels} have subsequently advanced NeRF's rendering quality by addressing challenges such as reducing aliasing artifacts~\cite{barron2021mip}, scaling to unbounded large-scale scenarios~\cite{barron2022mip}, and capturing specular reflections on glossy surfaces~\cite{verbin2022ref}.
Certain works~\cite{chen2022tensorf,fridovich2022plenoxels,mueller2022instant,kerbl20233d} have explored more effective representations of radiance fields. TensorsRF~\cite{chen2022tensorf} employs multiple compact low-rank tensor components, such as vectors and matrices, to represent the radiance field. Plenoxels~\cite{fridovich2022plenoxels} accelerates NeRF training by replacing MLPs with explicit plenoptic elements stored in sparse voxels and factorizing appearance through spherical-harmonic functions.
M\"uller \etal~\cite{mueller2022instant} achieved a substantial acceleration in rendering speed by employing a representation that combines trainable multi-resolution hash encodings (MHE) with shared shallow MLP networks. Kerbl \etal~\cite{kerbl20233d} introduce a novel volume rendering method utilizing 3D Gaussians to represent the radiance field and rendering images based on visibility-aware splatting of 3D Gaussians.

\subsection{Dynamic neural radiance fields} 
Neural fields \cite{xie2022neural} can be extended to represent dynamic scenes. On top of the \textit{canonical} scene representation, some methods~\cite{pumarola2020d, park2021nerfies, park2021hypernerf,yuan2021star} additionally model 4D deformation fields. Meanwhile, some other works learn a space-time correlated~\cite{kplanes_2023, li2020neural, attal2023hyperreel, liu2023robust} or decomposed~\cite{turki2023suds,wu2022d,yang2023emernerf} neural field to encode 4D scenes, achieving fine-grained reconstruction of geometry and appearance.
Some other methods decompose the scene into static and dynamic parts, and model each dynamic actor with dedicated neural fields. 
Neural Scene Graph~\cite{Ost_2021_CVPR} and Panoptic Neural Fields~\cite{KunduCVPR2022PNF} treat every dynamic object in the scene as a node, and synthesize photo-realistic RGB images by jointly rendering from both dynamic nodes and static background. UniSim\cite{yang2023unisim} employs the neural SDF representation to model dynamic scenes in driving scenarios, and renders them in a similar way as Neural Scene Graph~\cite{Ost_2021_CVPR}.

\subsection{Neural surface representation}
A fundamental challenge for NeRF and its variants is to accurately recover the underlying 3D surface from the implicit radiance field. Surfaces obtained by thresholding on the volume density of NeRF often exhibit noise~\cite{wang2021neus, yariv2021volume}. To address this, implicit surface representations like occupancy~\cite{niemeyer2020differentiable, oechsle2021unisurf} and signed distance functions (SDF)~\cite{wang2021neus, yariv2021volume, yu2022monosdf, sun2022neural, wang2022hf, zuo2023incremental, li2023neuralangelo, wang2023neus2} in grid maps are commonly integrated into neural volume rendering techniques.

NeuS~\cite{wang2021neus} introduces a neural SDF representation for surface reconstruction, proposing an unbiased weight function for the appearance composition process in volume rendering. Similarly, VolSDF~\cite{yariv2021volume} models scenes with a neural SDF and incorporates the SDF into the volume rendering process, advocating a sampling strategy for the viewing rays to bound opacity approximation error. Neuralangelo~\cite{li2023neuralangelo} improves surface reconstruction using multi-resolution hash encoding (MHE)~\cite{mueller2022instant} and SDF-based volume rendering~\cite{wang2021neus}. While these methods often deliver satisfactory surface reconstructions, their training is time-consuming, taking hours for a single scene.
Voxurf~\cite{wu2022voxurf} offers a faster surface reconstruction method through a two-stage training procedure, recovering the coarse shape first and refining details later. Wang \etal~\cite{wang2023neus2} expedite NeuS training to several minutes by predicting SDFs through a pipeline composed of MHE and shallow MLPs.

Many works also incorporate distances measured by LiDAR as auxiliary information to constrain the radiance field. For instance,~\cite{chang2023neural, wang2023neural} render depth by accumulating volume density and minimizing depth discrepancies between LiDAR and rendered depth during training. Rematas \etal~\cite{rematas2022urban} enforce empty space between the actual surface and the ray origin.

\subsection{LiDAR simulation} 
While simulators like CARLA~\cite{dosovitskiy2017carla} and AirSim~\cite{shah2018airsim} can simulate LiDAR data, they require expensive human annotations and suffer from a noticeable sim-to-real gap due to limited rendering quality. Generative model-based methods for LiDAR synthesis~\cite{caccia2019deep,zyrianov2022learning} offer an alternative but often lack control and produce distorted geometry~\cite{li2023pcgen}.
Learning-based approaches~\cite{li2023pcgen,fang2020augmented,manivasagam2020lidarsim} try to enhance realism by transferring real scan properties to simulations. For example, \cite{guillard2022learning} uses a RINet trained on RGB and real LiDAR data to enhance simulated scans. LiDARsim~\cite{manivasagam2020lidarsim} employs ray-surfel casting with explicit disk surfels for more accurate simulation.
Huang \etal~\cite{Huang2023nfl} have proposed Neural LiDAR Fields (NFL), combining neural fields with a physical LiDAR model for high-quality synthesis, but NFL is limited to static scenes and can produce noisy outputs due to its unconstrained volume density representation.
UniSim~\cite{yang2023unisim} constructs neural scene representations from realistic LiDAR and camera data, using SDF-based volume rendering to generate sensor measurements from novel viewpoints.
\section{Dynamic neural scene representation}

\paragraph{Problem statement.} 
Consider a set of LiDAR scans $\mathcal{X} = \{\mathbf{X}_t\}_{t=1}^T$ that have been compensated for ego-motion, along with tracked bounding boxes\footnote{We assume that these are are available.} for moving vehicles $\mathcal{B} = \{\mathbf{B}_t^v\}_{v=1}^{N}$, where $T$ represents the total number of LiDAR scans, and $N$ is the count of moving vehicles. Each scan $\mathbf{X}_t$ is composed of $n_t$ rays, each ray $\mathbf{r}$ is described by the tuple $(\origin, \dir, \zeta, \intensity, \pdrop)$, where $\origin$ and $\dir$ denote the ray's origin and direction, $\zeta$ and $\intensity$ represent range and intensity values, and $\pdrop \in \{0,1\}$ indicates whether the ray is dropped or not due to insufficient returned radiant power.

The goal is to reconstruct the scene with a static-dynamic decomposed neural representation, that can enable the rendering of LiDAR scan $\mathbf{X}_{\text{tgt}}$ from novel viewpoint $\mathbf{T}_{\text{tgt}}$. This setup also facilitates various object manipulations, including altering object trajectories, and inserting or removing objects from the scene. See~\cref{fig:main}.

\subsection{Neural scene decomposition} \label{sec: decomposition}
We leverage the inductive bias that driving scenes can be decomposed into a static background and $N$ rigidly moving dynamic components~\cite{huang2022dynamic,gojcic2021weakly}. Consequently, we establish $N+1$ neural fields. The neural field $\mathbf{F}_{\text{static}}$ is designated for the static component of the scene, capturing the unchanging background elements. Concurrently, the set of neural fields $\{\mathbf{F}^v\}_{v=1}^{N}$ is used to model the $N$ dynamic entities, specifically the vehicles in motion.

\vspace{-2mm}
\paragraph{Neural field for static background.} 
The static background is encoded into a neural field $\mathbf{F}_\text{static}: (\x, \dir) \mapsto (s, \intensity, \pdrop)$ that estimates the signed distance $s$, intensity $\intensity$, and ray drop probability $\pdrop \in [0,1]$ given the point coordinates $\x$ and the ray direction $\dir$. In practice, we first use a multi-resolution hash encoding (MRH)~\cite{mueller2022instant} to map each point to its positional feature $\posfeat \in \real^{32}$, and project the view direction onto the first 16 coefficients of the spherical harmonics basis, resulting in $\dirfeat$. Subsequently, we utilize three Multilayer Perceptrons (MLPs) to estimate the scene properties as follows:
\begin{equation}
(s, \geofeat) = f_s(\posfeat), \quad \intensity = f_{\intensity}(\rayfeat), \quad \pdrop = f_{\text{drop}}(\rayfeat).
\end{equation}
Here, $f_s, f_e,$ and $f_{\text{drop}}$ are three MLPs, $\rayfeat \in \mathbb{R}^{31}$ represents the ray feature and is constructed by concatenating the per-point geometric feature and the directional feature. The geometric feature is denoted as $\geofeat \in \mathbb{R}^{15}$. For implementation details please refer to the supplementary material. 

\vspace{-2mm}
\paragraph{Neural fields for dynamic vehicles.} 
LiDAR scans collected over time are often misaligned due to the motion of both the sensor and other objects in the scene. Despite applying ego-motion for aligning static background points, dynamic objects remain blurred along their trajectories. Our approach to constructing a dynamic neural scene representation is grounded in the assumption that each dynamic object only undergoes rigid motion. Therefore, we can first align them over time and reconstruct them in their \textit{canonical} coordinate frame, and then render them over time by reversing the alignment of the neural field.

Specifically, consider a dynamic vehicle $v$ 
occurring in LiDAR scans $\{\mathbf{X}^v_t\}_{t=1}^{T}$ along with the associated bounding boxes $\{\mathbf{B}^v_t \in \mathbb{R}^{3\times 8}\}_{t=1}^{T}$ in the world coordinate frame. Here each bounding box is defined by its eight corners, and the first bounding box $\mathbf{B}^v_1$ is considered as the \textit{canonical} box. We estimate the relative transformations $\{\mathbf{T}_t \in \text{SE}(3)\}_{t=2}^{T}$ between the remaining $T-1$ bounding boxes and the canonical one, expressed as $\mathbf{B}_1^v = \mathbf{T}_t \mathbf{B}_t^v$.\footnote{$\mathbf{T}\mathbf{B} = \mathbf{R}\mathbf{B} + \mathbf{t}$, with $\mathbf{R}$, $\mathbf{t}$ the rotation/translation components of $\mathbf{T}$.}. 
Subsequently, all LiDAR measurements on the object are transformed and accumulated in its canonical coordinate frame. The vehicle $v$ is then reconstructed in its canonical space, akin to the static background, using a neural field $\mathbf{F}^v$. To render the dynamic vehicle at timestamp $t$, the corresponding rigid transformation is applied to the queried rays. The dynamic neural field can thus be expressed as: $\mathbf{F}^v_t: (\mathbf{T}_{t}\x, \mathbf{T}_{t}\dir) \mapsto (s, \intensity, \pdrop)$. The rendering process for $\mathbf{F}^v$ is the same as for the static neural field $\mathbf{F}_{\text{static}}$.

\section{Neural rendering of the dynamic scene}
In this section, we present the methodology for rendering LiDAR scans from the neural scene representation. We begin by revisiting the density-based volume rendering formulation for active sensors~\cite{Huang2023nfl} in \cref{sec:vol_render_background}. Subsequently, we explore the extension of that formulation to the SDF-based neural scene representation in \cref{sec:sdf_vol_render}. Finally, we discuss in detail how to render LiDAR measurements from individual neural fields in~\cref{sec:dynamic_nfl_rendering} and how to compose the results from different neural fields in \cref{sec:neural_fields_composition}.

\subsection{Volume rendering for active sensor} \label{sec:vol_render_background}
LiDAR utilizes laser pulses to determine the distance to the nearest reflective surface, by analyzing the waveform profile of the returned radiant power. The radiant power $P(\zeta)$ from range $\zeta$ is the result of a convolution between the pulse power $P_e(t)$ and the impulse response $H(\zeta)$, defined as~\cite{hahner2021fog,hahner2022lidar,Huang2023nfl}:
\begin{equation}
    P(\zeta) = \int_0^{2\zeta/c} P_e(t) H(\zeta - \frac{ct}{2}) \; dt\;.
\label{eq:lidar}
\end{equation}
The impulse response $H(\zeta)$ is a product of the target and sensor impulse responses: $H(\zeta) = H_T(\zeta)\cdot H_S(\zeta)$, and the individual components are expressed as:
\begin{equation}
    H_T(\zeta) = \frac{\reflectance}{\pi} \cos(\theta) \delta(\zeta - \bar{\zeta})\;, \quad  H_s(\zeta) = T^2(\zeta) \frac{A_e}{\zeta^2}\;,
\label{eq:ht}
\end{equation}
where $\reflectance$ represents the surface reflectance, $\theta$ denotes incidence angle, $\bar{\zeta}$ is the ground truth distance to the nearest reflective surface, $T(\zeta)$ and $A_e$ describe the transmittance at range $\zeta$ and effective sensor area, respectively. Due to the discontinuity introduced by the indicator function $\delta(\zeta - \bar{\zeta})$, ~\cref{eq:lidar} is non-differentiable and is thus not suitable for solving the inverse problem. NFL~\cite{Huang2023nfl} solves it by extending it into a probabilistic formulation given by:
\begin{equation}
P(\zeta) = C \cdot \frac{T^2(\zeta) \cdot \density_\zeta  \reflectance_\zeta}{\zeta^2} \cos(\theta)\;.
\label{eq:radiance}
\end{equation}
Here, $C$ accounts for the constant values, and $\sigma_\zeta$ represents the density at range $\zeta$. The radiant power can be reconstructed using the volume rendering equation:
\begin{equation}
      P
      =\!\sum_{j=1}^N \int_{\zeta_j}^{\zeta_{j+1}}\!\!C \frac{T^2({\zeta}) \cdot \density_\zeta \reflectance_\zeta}{\zeta^2} \cos(\theta_j) \; d\zeta
      =\!\sum_{j=1}^N w_j \reflectance_{\zeta_j}',
\label{eq:radiant_inter}
\end{equation}
where the weights $w_j = 2 \opacity_{\zeta_j} \cdot\prod_{i=1}^{j-1}(1 - 2 \opacity_{\zeta_i}).$
~Here $\alpha_{\zeta_j}$ is the discrete opacity at range $\zeta_j$. Please refer to~\cite{Huang2023nfl} for more details.

\subsection{SDF-based volume rendering for active sensor} \label{sec:sdf_vol_render}
A neural scene representation based on probabilistic density often results in surfaces with noticeable noise due to insufficient surface regularization~\cite{wang2021neus}. To address this, we opt for a signed distance-based scene representation and establish the LiDAR volume rendering formulation within that framework. Building upon SDF-based volume rendering for passive sensors~\cite{wang2021neus}, we compute the opaque density $\tilde{\density}_{\zeta_i}$ as follows:
\begin{equation}
\tilde{\density}_{\zeta_i} = \max\left(\frac{-\frac{{\rm d}\Phi_s}{{\rm d} \zeta_i}(f(\zeta_i))}{\Phi_s(f(\zeta_i))},0\right),
\label{eq:sigmoid_density}
\end{equation}
where $\Phi_s(\cdot)$ represents the Sigmoid function, $f(\zeta)$ evaluates the signed distance to the surface at range $\zeta$ along the ray $\ray$. 
Next, we substitute the density $\density$ in \cref{eq:radiant_inter} with the opaque density from \cref{eq:sigmoid_density} and re-evaluate the radiant power and weights as:
\begin{equation}
      P
      =\!\sum_{j=1}^N \trans^2_{\zeta_j} \tilde{\alpha}_{\zeta_j} \reflectance_{\zeta_j}',\quad \tilde{w}_j = 2 \tilde{\opacity}_{\zeta_j} \cdot\prod_{i=1}^{j-1}(1 - 2 \tilde{\opacity}_{\zeta_i})\;.
\end{equation}
In this context, $\tilde{\alpha}_{\zeta_j}$ is computed as:
\begin{equation}
    \tilde{\alpha}_{\zeta_j} = \max\left(\!\frac{{\Phi_s(f(\zeta_j))}^2 -{\Phi_s(f(\zeta_{j+1}))}^2}{{2\Phi_s(f(\zeta_j))}^2},0\right).
    \label{eq:new_weights}
\end{equation}
Please refer to the supplementary material for more details.

\subsection{Volume rendering for LiDAR measurements}\label{sec:dynamic_nfl_rendering}
To render the LiDAR measurements from a single neural field, we employ the hierarchical sampling technique~\cite{wang2021neus} to sample a total of $N_s= N_u + N_i$ points along each ray, where $N_u$ points are uniformly sampled, and $N_i$ points are probabilistically sampled based on the weights along the ray for denser sampling in the proximity of the surface. Subsequently, we compute the weights for the $N_s$ points following~\cref{eq:new_weights}. The rendering of range, intensity, and ray drop for each ray can be expressed through volume rendering as follows: $y_\text{est} = \sum_{j=1}^{N_s} w_j y_j$, where $y \in \{\zeta, \intensity, \pdrop\}$.

\subsection{Neural rendering for multiple fields}\label{sec:neural_fields_composition}
Our full neural scene representation comprises $N+1$ neural fields as discussed in Sec.~\ref{sec: decomposition}. Rendering from all these fields for each ray during inference is computationally intensive. To address this, we implement a two-stage method. In the first stage, we identify $k \geq 0$ dynamic fields that are likely to intersect with a given ray, plus the static background. The second stage renders LiDAR measurements from these selected fields individually and then integrates them into a unified set of measurements.

\vspace{-3mm}
\paragraph{Ray intersection test.}
As outlined in~\cref{sec: decomposition}, each dynamic neural field is reconstructed in its unique canonical space, defined by a corresponding canonical box. To determine neural fields that intersect a given ray at inference time, we begin by estimating the transformations $\{\mathbf{T}_t^v\}_{v=1}^N$, which convert coordinates from the world framework to each vehicle's canonical space at timestamp $t$. These transformations are determined by interpolating the training set transformations using spherical linear interpolation (SLERP)~\cite{10.1145/325334.325242}. Then, we apply the transformations to the queried ray and run intersection tests with the canonical boxes of the fields. 

\vspace{-3mm}
\paragraph{Neural rendering from multiple neural fields.}
 After identifying the $k+1$ neural fields that potentially intersect with a ray, we run volume rendering on each field separately, yielding $k+1$ distinct sets of LiDAR measurements. Next, we evaluate the ray drop probabilities across these fields. A ray is deemed \textit{dropped} if all neural fields indicate a drop probability $\pdrop > 0.5$. For rays not classified as dropped, we sort the estimated ranges in ascending order and select the nearest one as our final range prediction. The corresponding intensity value at the closest range is extracted from the same neural field.
\section{Neural scene optimisation} \label{sec:optmisation}
Given the set of LiDAR scans and the associated tracked bounding boxes of the dynamic vehicles, we optimise our neural scene representation by minimising the loss:
\begin{equation}
    \mathcal{L} = w_{\zeta} \mathcal{L}_{\zeta} +  w_{s} \mathcal{L}_{s} + w_{\text{eik}} \mathcal{L}_{\text{eik}} + w_{\intensity} \mathcal{L}_{\intensity} + w_{\text{drop}} \mathcal{L}_{\text{drop}},
    \label{loss}
\end{equation}
where the $w_{*}$ denote the weights of the individual loss terms $\mathcal{L}_*$, explained below.

\vspace{-2mm}
\paragraph{Range reconstruction loss.}
For range estimation, we employ an L1 loss, defined as: $\mathcal{L}_{\zeta} = \frac{1}{|\mathcal{R}|}\sum_{\ray \in \mathcal{R}}|\zeta_{est} -\zeta_{gt}|$, where $\mathcal{R}$ denotes the set of LiDAR rays, $\zeta_{est}$ and $\zeta_{gt}$ are the estimated and actual ranges, respectively. 

\vspace{-2mm}
\paragraph{Surface point SDF regularisation.} \label{sec:surfacesdf}
Acknowledging that LiDAR points mostly come from the actual surface, we introduce an additional SDF regularisation term $\mathcal{L}_{s}$ that penalizes non-zero SDF values at surface points: $\mathcal{L}_{s} = \frac{1}{|\mathcal{P}|}\sum_{\mathbf{p} \in \mathcal{P}}|s(\mathbf{p})|$. Here $\mathcal{P}$ denotes the set of surface points and $s({\mathbf{p}})$ represents the SDF value of the point $\mathbf{p}$.

\vspace{-2mm}
\paragraph{Eikonal constraint.}
Following~\cite{icml2020_2086}, we utilize the Eikonal loss $\leik$, to regularize the SDF level set. This ensures the gradient norm of the SDF is approximately one at any queried point. The loss is computed as: $\leik = \frac{1}{|\mathcal{Z}|} \sum_{\mathbf{p} \in \mathcal{Z}}( \| \nabla s(\mathbf{p}) \|_2 - 1)^2$, where $\mathcal{Z}$ is the set of all the sampled points. To stabilise the training procedure, we adopt a numerical approach~\cite{li2023neuralangelo} to compute $\nabla s(\pos)$ as: 
\begin{equation}
    \nabla s(\pos) = \frac{s \left( \pos + \boldsymbol{\epsilon} \right) - s \left(\pos - \boldsymbol{\epsilon} \right)}{2 \epsilon} \;,
    \label{eqn:central_diff_normal}
\end{equation}
where the numerical step size $\epsilon$ is set to be $10^{-3}$ meters.

\vspace{-2mm}
\paragraph{Intensity Loss.}
For intensity reconstruction, we apply the L2 loss, defined as: $\mathcal{L}_{\intensity} = \frac{1}{|\mathcal{R}|}\sum_{\ray \in \mathcal{R}}(\intensity_{est} -\intensity_{gt})^2.$

\vspace{-2mm}
\paragraph{Ray drop loss.}
We follow~\cite{Huang2023nfl} to supervise the ray drop estimation with a combination of a binary cross entropy loss $\mathcal{L}_{bce}$ and a Lovasz loss $\mathcal{L}_{ls}$ \cite{berman2018lovasz} as:
\begin{equation}
     \mathcal{L}_{\text{drop}} = \frac{1}{|\mathcal{R}|} \sum_{\ray \in \mathcal{R}} \left(\mathcal{L}_{bce}(p_{d, est}, {p_{d, gt}}) + \mathcal{L}_{ls}(p_{d, est}, {p_{d, gt}}) \right)\;.
     \label{eq:raydrop_loss}
\end{equation}
Notably, in the context of dynamic neural field training, we include \emph{all} LiDAR rays that intersect with the objects' bounding boxes in a scene. A ray is classified as \textit{dropped} either if it is labeled as such in the dataset or if it does not intersect with the actual surface of a dynamic vehicle (\eg, rays close to and parallel with the surfaces). This approach enhances the accuracy and realism of the reconstructed dynamic neural fields, improving the rendering fidelity at inference time. 
\section{Experiments}
\begin{figure*}[t]
  \centering
   \includegraphics[width=1\textwidth]{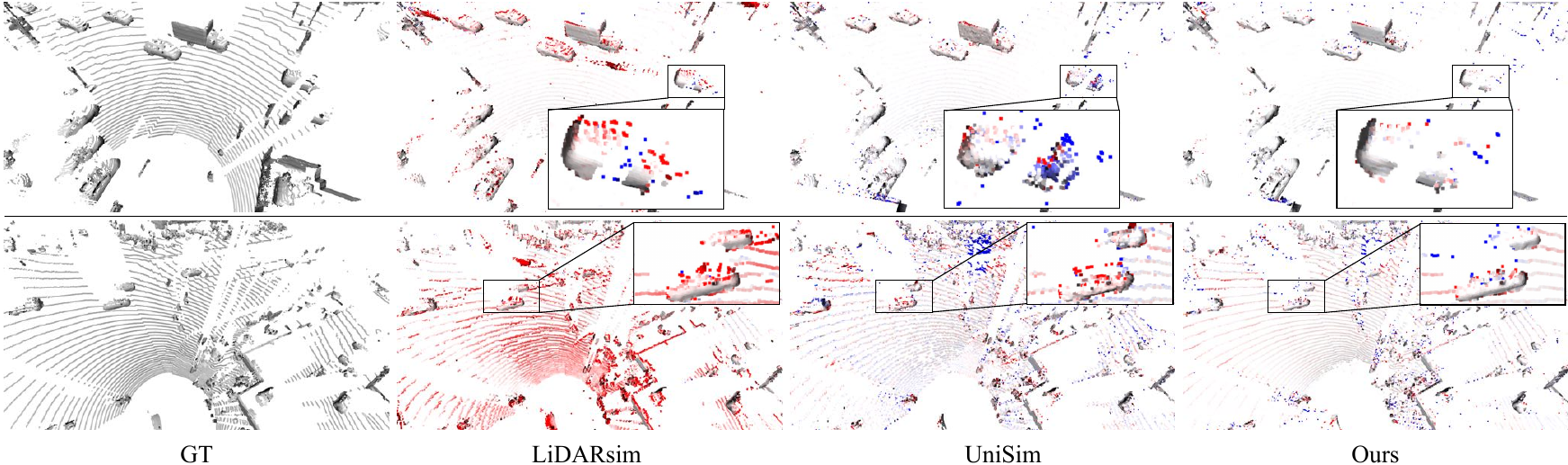}
   \vspace{-5mm}
   \caption{Qualitative comparison of range estimation on \textit{Waymo Dynamic} dataset. Dynamic vehicles are zoomed in, and points are color-coded by range errors~(-100 \bwr~100 cm).
   }
   \label{fig:errormap_dynamic}
\end{figure*}

\subsection{Datasets and evaluation protocol}\label{sec:datasets}

\paragraph{Real-world dynamic scenes.} 
We construct the \textit{Waymo Dynamic} dataset by selecting four representative scenes from Waymo Open~\cite{sun2020scalability}, with multiple moving vehicles inside. These scenes are comprised of 50 consecutive frames. For evaluation purposes, every 5\textsuperscript{th} frame is held out for testing, the remaining 40 frames are used for training.

\vspace{-2mm}
\paragraph{Real-world static scenes.}
We also evaluate our method on four static scenes as introduced in~\cite{Huang2023nfl}. There are two settings, \textit{Waymo Interp} applies the same evaluation protocol as \textit{Waymo Dynamic}, while \textit{Waymo NVS} employs a dedicated closed-loop evaluation to validate the real novel view synthesis performance. Please refer to NFL~\cite{Huang2023nfl} for more details about this setting.

\vspace{-2mm}
\paragraph{Synthetic static scenes.}  
\textit{TownClean} and \textit{TownReal} are synthetic static scenes introduced in NFL~\cite{Huang2023nfl}. They consist of 50 simulated LiDAR scans of urban street environments, using idealised and diverging beams, respectively.

\vspace{-2mm}
\paragraph{Evaluation metrics.}\label{sec:metrics}
To evaluate the LiDAR range accuracy, we employ a suite of four metrics: mean absolute errors~(MAE [cm]), median absolute errors~(MedAE [cm]), Chamfer distance~(CD [cm]) and MedAE for dynamic vehicles~(MedAE Dyn [cm]). For intensity evaluation, We report root mean square error~(RMSE).
In addition to our primary evaluations, we assess the re-simulated LiDAR scans' realism through two auxiliary tasks: object detection and semantic segmentation. For object detection, we calculate the \textit{detection agreement}~\cite{manivasagam2020lidarsim}, both for all vehicles (Agg.~[\%]) and specifically for dynamic vehicles (Agg.$\;$Dyn.~[\%]). Regarding semantic segmentation, we report recall, precision, and intersection-over-union (IoU~[\%]). It is important to note that the predictions on the original LiDAR scans serve as our \textit{ground truth}, against which we compare the results obtained from the re-simulated scans.

\vspace{-2mm}
\paragraph{Baseline methods.}
Regarding LiDAR simulation on static scenes, NFL~\cite{Huang2023nfl} and LiDARsim~\cite{manivasagam2020lidarsim} are the two closest baselines to compare to. Additionally, we include i-NGP~\cite{mueller2022instant}, DS-NeRF~\cite{kangle2021dsnerf}, and URF~\cite{rematas2022urban} for comparison. As for simulation on dynamic scenes, we compare to LiDARsim~\cite{manivasagam2020lidarsim} and UniSim~\cite{yang2023unisim}.\footnote{We re-implement both LiDARsim and UniSim, since they are not open-sourced.} Please refer to the supplementary material for implementation details.

\begin{table}[t]
    \setlength{\tabcolsep}{4pt}
    \renewcommand{\arraystretch}{1.2}
	\centering
	\resizebox{\columnwidth}{!}{
    \begin{tabular}{l|ccccc}
    \toprule
    Method  & MAE $\downarrow$ &  MedAE $\downarrow$ & CD $\downarrow$ & MedAE Dyn $\downarrow$ & Intensity RMSE $\downarrow$\\
    \midrule
    LiDARsim~\cite{manivasagam2020lidarsim} & 170.1 & 11.5 & 31.1 &  16.0  & 0.10\\
    Unisim~\cite{yang2023unisim} & 35.6 & 6.1 & 14.3 &14.3 & \textbf{0.05}\\
    Ours~ & \textbf{30.8} & \textbf{3.0} & \textbf{10.9} &\textbf{8.5} & \textbf{0.05}\\
    \bottomrule
    \end{tabular}
    }
    \vspace{-2mm}
	\caption{Evaluation of LiDAR NVS on \textit{Waymo Dynamic} dataset.}
	\label{tab:waymodynamic}
\end{table}
\begin{table}[t]
    \setlength{\tabcolsep}{4pt}
    \renewcommand{\arraystretch}{1.2}
	\centering
	\resizebox{\columnwidth}{!}{
    \begin{tabular}{l|ccc|ccc|ccc|ccc}
    \toprule
    & \multicolumn{3}{c|}{TownClean}& \multicolumn{3}{c|}{TownReal} & \multicolumn{3}{c|}{Waymo interp.} & \multicolumn{3}{c}{Waymo NVS} \\
    Method  & MAE $\downarrow$ &  MedAE $\downarrow$ & CD $\downarrow$& MAE $\downarrow$ &  MedAE $\downarrow$ & CD $\downarrow$ & MAE $\downarrow$ &  MedAE $\downarrow$ & CD $\downarrow$ & MAE $\downarrow$ &  MedAE $\downarrow$ & CD $\downarrow$\\
    \midrule
    i-NGP~\cite{mueller2022instant} &42.2 &4.1 & 17.4 & 49.8 & 4.8 & 19.9 & \textbf{26.4} & 5.5 & \textbf{11.6} & \underline{30.4} & 7.3 & 15.3\\
    DS-NeRF~\cite{kangle2021dsnerf} &41.7 & 3.9 &16.6 & 48.9 & 4.4 & 18.8 & \underline{28.2} & 6.3 & 14.5 & 30.4 & 7.2 & 16.8 \\
    URF~\cite{rematas2022urban} &43.3&4.2&16.8& 52.1 & 5.1 & 20.7 & 28.2 & 5.4 & 12.9 & 43.1 & 10.0 & 21.2 \\
    LiDARsim~\cite{manivasagam2020lidarsim} &159.6&\underline{0.8}&23.5& 162.8 & 3.8 & 27.4 & 116.3 & 15.2 & 27.6 & 160.2 & 16.2 & 34.7 \\
    NFL\cite{Huang2023nfl}  &\underline{32.0}&2.3&\underline{9.0}& \underline{39.2} & \underline{3.0} & \underline{11.5} & 30.8 & \underline{5.1} & \underline{12.1}& 32.6 & \underline{5.5} & \underline{13.2}  \\
    Ours & \textbf{26.7} & \textbf{0.7} & \textbf{6.7} &\textbf{33.9}&\textbf{2.1}&\textbf{10.4}& 28.3 & \textbf{4.7} & 12.5 & \textbf{28.6} & \textbf{4.9} & \textbf{13.0} \\
    \bottomrule
    \end{tabular}
}
    \vspace{-2mm}
	\caption{Evaluation of LiDAR NVS on static scenes.}
	\label{tab:waymostatic}
\end{table}
\begin{figure}[t]
  \centering
   \includegraphics[width=1\columnwidth]{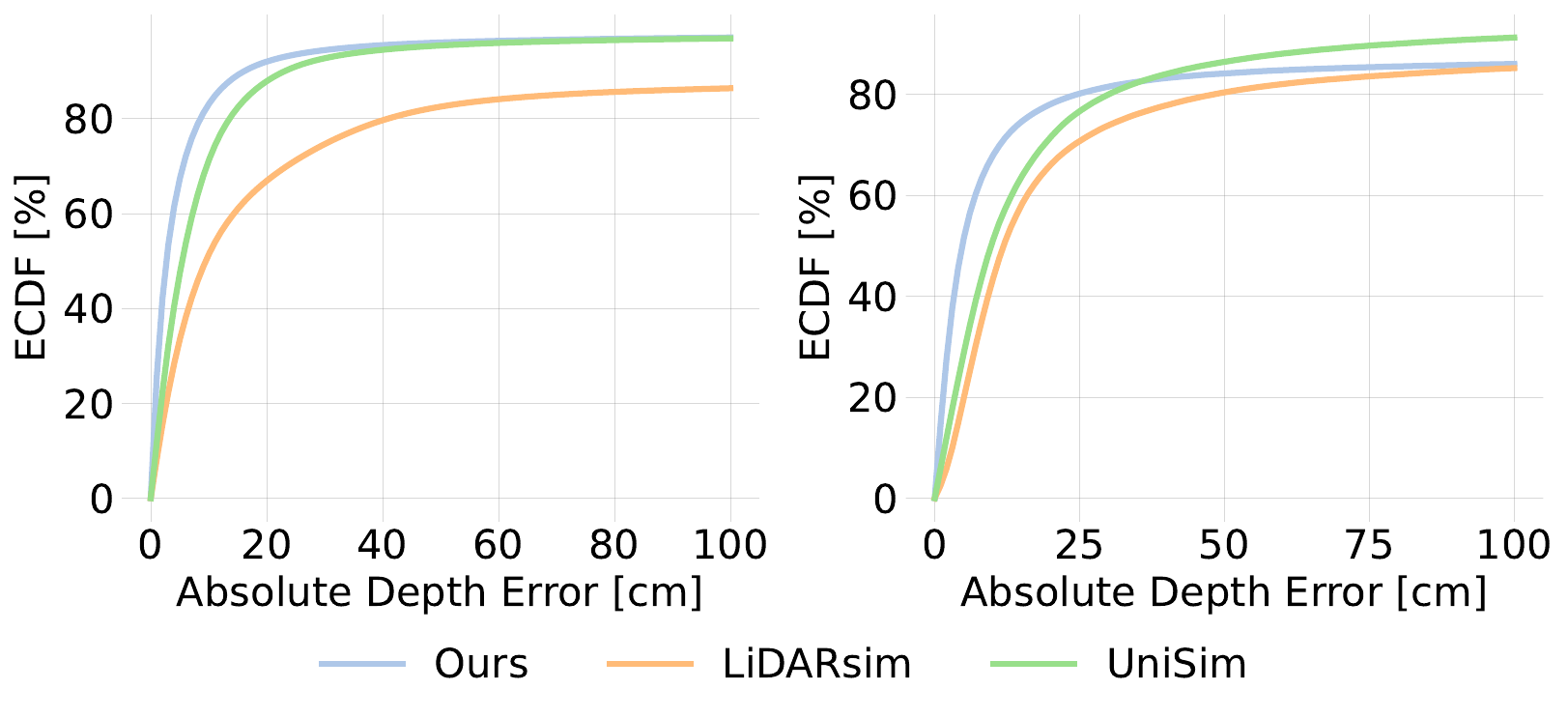}
   \vspace{-5mm}
        \caption{ECDF plots showcasing range errors across all the points (left) and specifically for points on dynamic vehicles (right). Our composition of neural fields outperforms LiDARsim~\cite{manivasagam2020lidarsim} and UniSim~\cite{yang2023unisim}, especially when it comes to dynamic vehicles.}
   \label{fig:ecdf}
\end{figure}
\begin{figure}[t]
    \centering
        \includegraphics[width=1\linewidth]{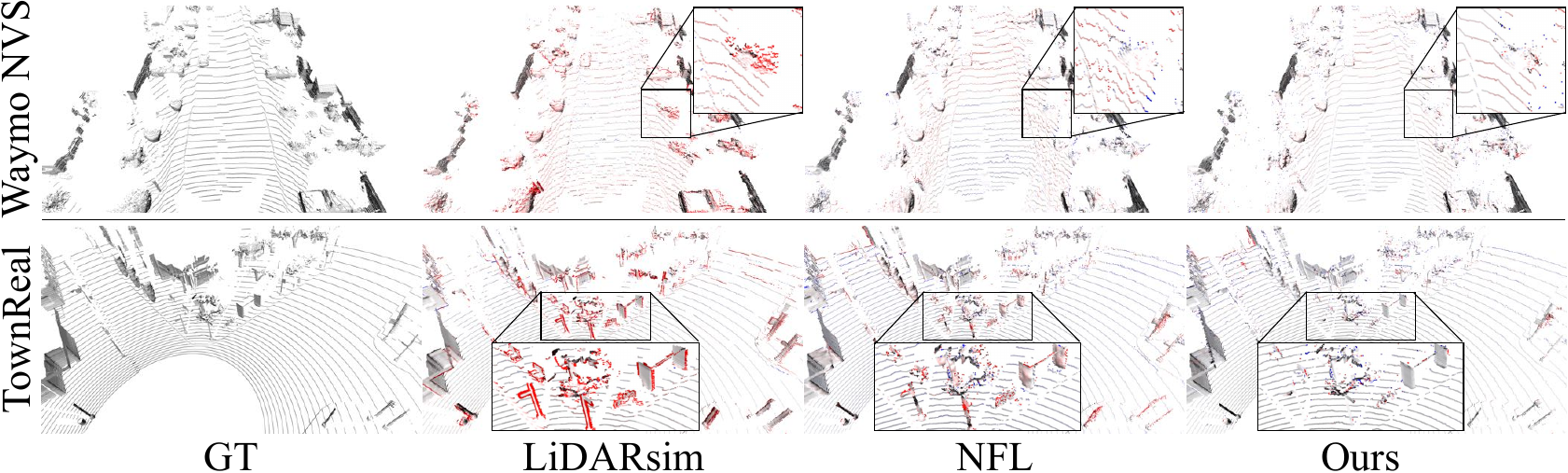}
        \vspace{-5mm}
        \caption{Qualitative results of range estimation. Regions with gross errors (-100 \bwr~100 cm) are highlighted.
        }
    \label{fig:error_map}
\end{figure}
\subsection{LiDAR novel view synthesis evaluation} \label{sec:lidar_eval}
\paragraph{LiDAR NVS in dynamic scenes.}

Quantitative comparisons with baseline methods are detailed in~\cref{tab:waymodynamic}. \method reconstructs more accurate range than LiDARsim~\cite{manivasagam2020lidarsim} and UniSim~\cite{yang2023unisim}. The improvement is largely due to our SDF-based neural scene representation, which incorporates the physical aspects of LiDAR sensing. Additionally, our method employs a ray drop test when rendering multiple neural fields, leading to a more accurate reconstruction of dynamic vehicles, as evidenced in~\cref{fig:errormap_dynamic} and further supported by the errors shown in~\cref{fig:ecdf}.

\vspace{-2mm}
\paragraph{LiDAR NVS in static scenes.}
In addition to dynamic scenes, we evaluate \method against baseline methods in static scenarios, \cref{tab:waymostatic} and~\cref{fig:error_map}. \method excels in reconstructing crisp geometry. A key observation is its superior performance on planar regions (\eg, the ground shown in~\cref{fig:error_map}), especially when compared to NFL~\cite{Huang2023nfl}, which also uses a neural field for surface representation. This improvement is largely due to the enhanced surface regularization provided by SDF-based surface modeling.

\begin{table}[t]
    \setlength{\tabcolsep}{4pt}
    \renewcommand{\arraystretch}{1.2}
	\centering
	\resizebox{0.75\columnwidth}{!}{
    \small
    \begin{tabular}{l|ccc}
    \toprule
    Datasets  & MAE $\downarrow$ &  MedAE $\downarrow$ & CD $\downarrow$ \\
    \midrule
    TownClean~ & 26.7(\textcolor{green}{-1.5}) & 0.7(\textcolor{green}{-0.2}) & 6.7(\textcolor{green}{-0.5})\\
    Waymo Interp~ & 28.3 (\textcolor{red}{0.1}) & 4.7 (\textcolor{green}{-0.2}) & 12.5 (\textcolor{green}{-0.1})\\
    Waymo Dynamic~ & 30.8 (\textcolor{green}{-0.3}) & 3.0 (\textcolor{green}{-0.2}) & 10.9 (\textcolor{green}{-0.3})\\
    \bottomrule
    \end{tabular}
    }
    \vspace{-2mm}
	\caption{Ablation study of volume rendering for active sensing.}
	\label{tab:active_sensing}
\end{table}
\begin{table}[t]
    \setlength{\tabcolsep}{4pt}
    \renewcommand{\arraystretch}{1.2}
	\centering
	\resizebox{0.75\columnwidth}{!}{
    \small
    \begin{tabular}{l|ccc}
    \toprule
    Datasets  & MAE $\downarrow$ &  MedAE $\downarrow$ & CD $\downarrow$ \\
    \midrule
    TownReal~ & 33.9(\textcolor{green}{-3.3}) & 2.1(\textcolor{green}{-0.0}) & 10.4(\textcolor{green}{-1.2})\\
    Waymo Interp~ & 28.3 (\textcolor{green}{-0.3}) & 4.7 (\textcolor{green}{-0.1}) & 12.5 (\textcolor{green}{-0.3})\\
    \bottomrule
    \end{tabular}
    }
    \vspace{-2mm}
	\caption{Ablation study of the surface points' SDF regularisation.}
	\label{tab:surface_sdf}
\end{table}
\begin{figure}[t]
  \centering
   \includegraphics[width=1\linewidth]{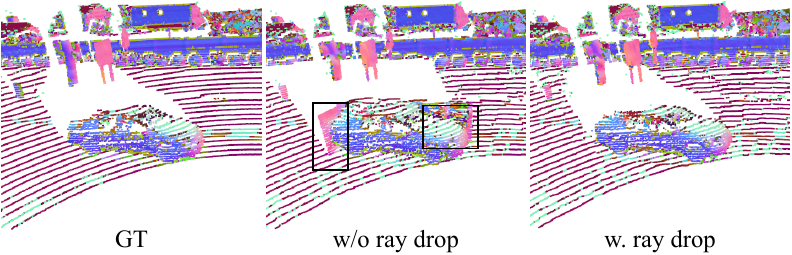}
   \vspace{-5mm}
   \caption{
   Qualitative results on \textit{Waymo Dynamic} dataset. Our model equipped with a ray drop module effectively composites multiple neural fields, re-simulating LiDAR scans of high quality.
   }
   \label{fig:ablation_raydrop}
\end{figure}

\begin{figure}[t]
    \centering
        \includegraphics[width=1\linewidth]{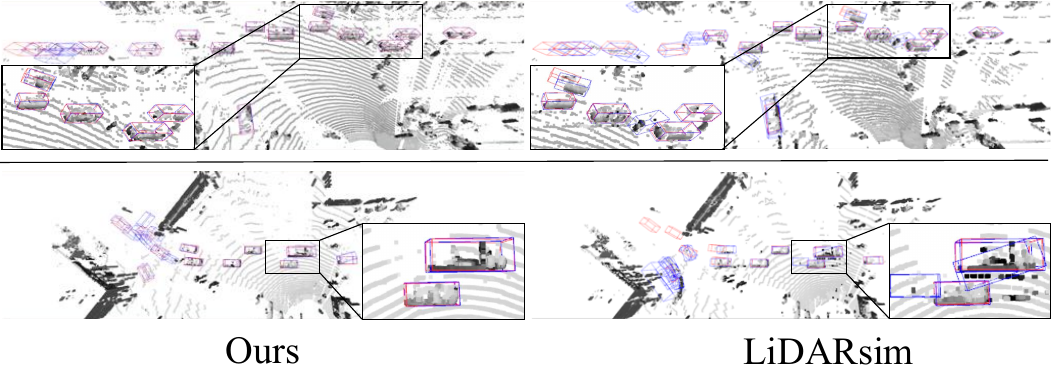}
        \vspace{-3mm}
        \caption{Object detection results on \textit{Waymo Dynamic} dataset. The ground truth and predicted bounding boxes are marked in \textcolor{red}{red} and \textcolor{blue}{blue}, respectively.}  
    \label{fig:detection}
\end{figure}
\begin{figure*}[t]
    \centering
        \includegraphics[width=1.0\textwidth]
        {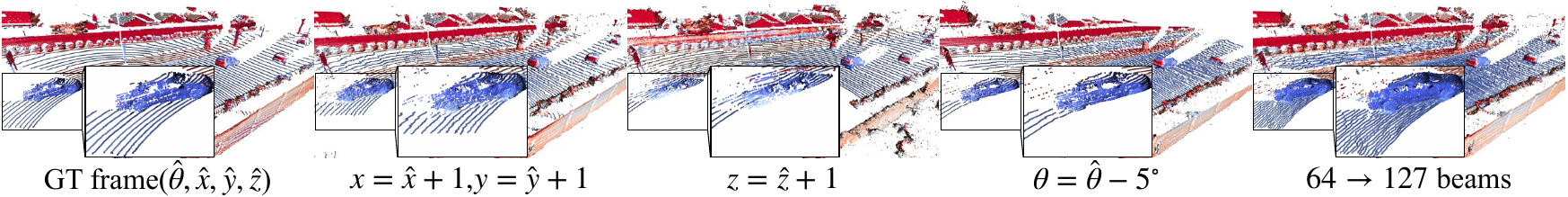}
        \vspace{-6mm}
        \caption{LiDAR novel view synthesis by changing sensor elevation angle~($\theta$), poses~($x,y,z$) and number of beams on \textit{Waymo Dynamic} dataset. The points are color-coded by the intensity values (0 \bwr~0.25).}
    \label{fig:lidar_nvs}
\end{figure*}
\begin{table}[t]
	\centering
	\resizebox{0.9\columnwidth}{!}{
		\begin{tabular}{@{}lcccccccc@{}}
			\toprule
             & \multicolumn{1}{c}{GT} & \multicolumn{3}{c}{Ours} & \multicolumn{3}{c}{LiDARSim\cite{manivasagam2020lidarsim}} \\
			  \cmidrule(r){2-2}\cmidrule(r){3-5} \cmidrule(l){6-8}
			Threshold & AP~$\uparrow$ & \multicolumn{1}{c}{AP~$\uparrow$}& \multicolumn{1}{c}{Agg.~$\uparrow$}& \multicolumn{1}{c}{Dyn. Agg.~$\uparrow$} & \multicolumn{1}{c}{AP~$\uparrow$} & \multicolumn{1}{c}{Agg.~$\uparrow$}& \multicolumn{1}{c}{Agg.~Dyn.~$\uparrow$} \\
			\midrule
			IoU$>$0.7 &0.85  &0.86 & \textbf{0.77}& \textbf{0.71}& \textbf{0.90} & 0.76 & 0.68\\
			IoU$>$0.5 &\textbf{0.98}  & 0.96 & \textbf{0.87}& \textbf{0.76}& 0.95 & 0.86& \textbf{0.76} \\
			\bottomrule
		\end{tabular}
	}
    \vspace{-2mm}
	\caption{Object detection results on \textit{Waymo Dyanmic} datasets.}
	\label{tab:detection}
\end{table}
\begin{table}[t]
\setlength{\tabcolsep}{4pt}
\renewcommand{\arraystretch}{1.2}
\centering
\resizebox{0.99\columnwidth}{!}{
\begin{tabular}{l|ccc|ccc}
\toprule
& \multicolumn{3}{c|}{Vehicle} & \multicolumn{3}{c}{Background} \\
Method & Recall $\uparrow$ & Precision $\uparrow$ & IoU $\uparrow$ & Recall $\uparrow$ & Precision $\uparrow$ & IoU $\uparrow$ \\
\midrule
i-NGP~\cite{mueller2022instant} & 91.8 & 83.6 & 78.1 & 97.9 & 99.2 & 97.1\\
DS-NeRF~\cite{kangle2021dsnerf} & 89.3 & 84.8 & 77.3 & 98.1 & 98.8 & 97.0\\
URF~\cite{rematas2021urban} & 86.9 & 79.8 & 72.0 & 97.7 & 98.5 & 96.2\\
Lidarsim~\cite{manivasagam2020lidarsim} & 89.6 & 68.9 & 64.0 & 94.5 & 98.9 & 93.5\\
NFL~\cite{Huang2023nfl}& \textbf{94.5} & 84.8 & 80.9 & 97.8 & \textbf{99.4} & \textbf{97.3}\\
Ours & 90.5 & \textbf{88.4} & \textbf{81.1} & \textbf{98.5} & 98.7 & \textbf{97.3}\\
\bottomrule
\end{tabular}
}
\vspace{-2mm}
\caption{Semantic segmentation results on \textit{Waymo NVS} dataset.}
\label{tab:sem_seg_nvs_ours}
\end{table}
\begin{figure}[t]
    \includegraphics[width=1.0\linewidth]{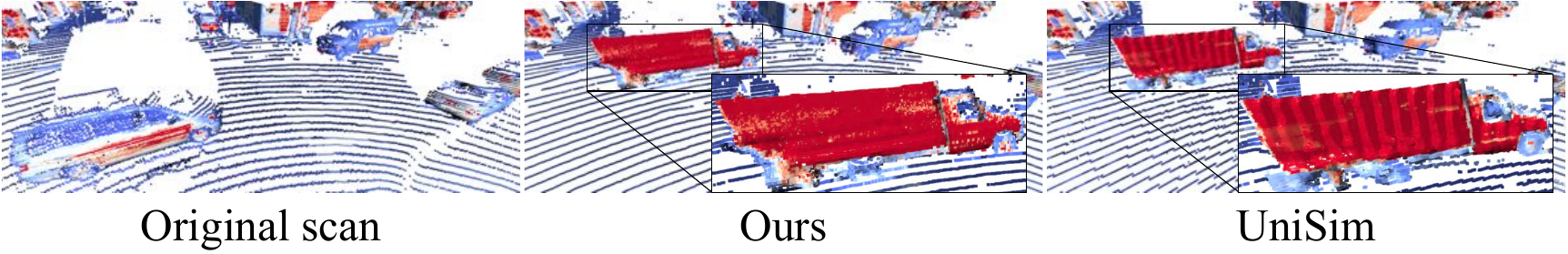}
    \vspace{-6mm}
    \caption{
    Qualitative results of object removal and insertion. \method seamlessly inserts the neural asset (truck) into a new scene attributed to our superior compositional rendering scheme. In contrast, UniSim~\cite{yang2023unisim} struggles to accurately model geometry.
    }
    \label{fig:vehicle_insertion}
\end{figure}
\begin{figure}[t]
    \includegraphics[width=1.0\columnwidth]{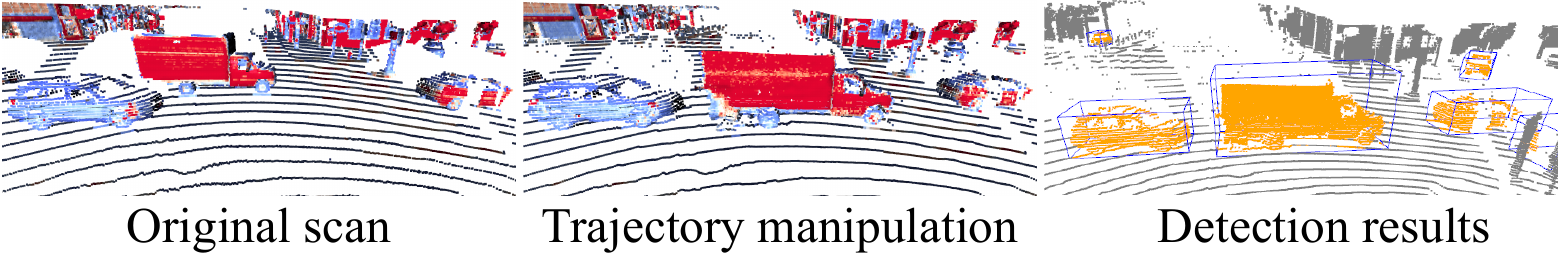}
    \vspace{-6mm}
    \caption{Qualitative results of object trajectory manipulation. The truck can be successfully detected after manipulation, indicating high-realism LiDAR re-simulation achieved by \method.}
    \label{fig:traj}
\end{figure}
\begin{figure}[t]
   \centering
   \includegraphics[width=1\linewidth]{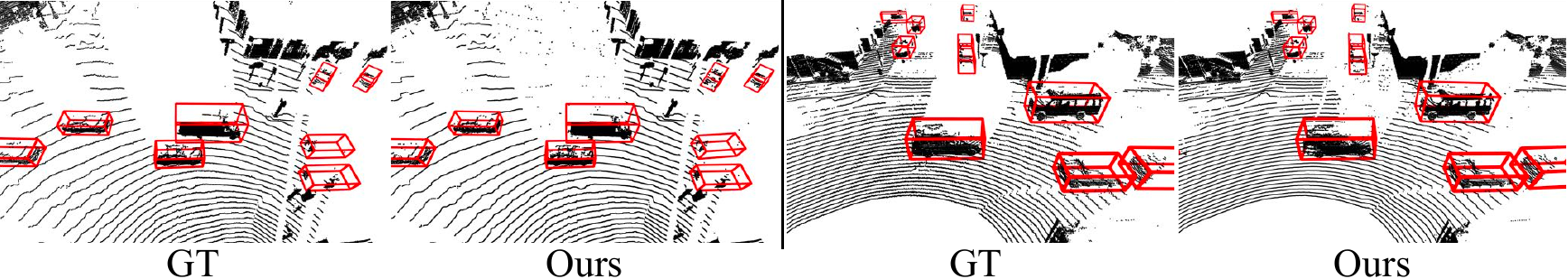}
   \vspace{-6mm}
   \caption{
   Object detections on noisy re-simulated LiDAR scans. 
   }
   \label{fig:noisy_points}
\end{figure}

\subsection{Ablation study}
\paragraph{SDF-based volume rendering for active sensing.}
We begin by assessing the efficacy of our SDF-based volume rendering for active sensors, results are shown in~\cref{tab:active_sensing}. When compared to our baseline that uses the SDF-based volume rendering for passive sensing, \method demonstrates enhanced performance in both synthetic (\textit{TownClean}) and real-world (\textit{Waymo Interp} and \textit{Waymo Dynamic}) datasets, indicating the importance of considering the physical LiDAR sensing process when addressing the inverse problem.

\vspace{-3mm}
\paragraph{Neural fields composition.} 
To validate the efficacy of our two-stage neural field composition approach, we compare it with an alternative approach utilized in UniSim~\cite{yang2023unisim}. The results are shown in~\cref{tab:waymodynamic}. UniSim~\cite{yang2023unisim} blends different neural fields by sampling points from all intersected neural fields, followed by a single evaluation of volume rendering to produce the final LiDAR scan. In contrast, our method independently renders from each intersecting neural field first, and then combines these measurements into a final output using a ray drop test (\cf~\cref{fig:ablation_raydrop}). This approach leads to markedly improved geometry reconstruction, exemplified by our method halving the Median Absolute Error (MedAE) across all points. This enhancement is even more evident when focusing solely on points related to dynamic vehicles (\cf~\cref{fig:ecdf}).
\vspace{-3mm}
\paragraph{Surface point SDF constraint.}
We examine the importance of constraining the SDF at surface points, \cf~\cref{sec:optmisation}, with the \textit{Town Real} and \textit{Waymo Interp} datasets. The results shown in \cref{tab:surface_sdf} suggest that our method yields improved geometry reconstruction by explicitly enforcing SDF values near zero at the LiDAR points.

\subsection{Auxiliary task evaluations} 
\label{sec:downstream}
To assess the fidelity of our neural re-simulation and gauge the domain gap between re-simulated and real scans, we evaluate their applicability in two downstream tasks: object detection and semantic segmentation.

\vspace{-3mm}
\paragraph{Object detection.}
We run the pre-trained FSDv2~\cite{fan2023fsdv2} object detector on re-simulated LiDAR scans of the \textit{Waymo Dynamic} dataset. Our results are compared against those from LiDARsim~\cite{manivasagam2020lidarsim}, with the findings detailed in~\cref{tab:detection} and~\cref{fig:detection}. In summary, detections in scans generated with~\method exhibit better agreement with those in real LiDAR scans. In~\cref{fig:noisy_points} we additionally show detection results in the noisy background. The detections in the synthetic scans almost perfectly match those in the original scans. This indicates a high fidelity of our re-simulations, with a low domain gap to actual scans.

\vspace{-3mm}
\paragraph{Semantic segmentation.}
We segment synthetic scans into semantic classes with the pre-trained SPVNAS model~\cite{tang2020searching} and show results in~\cref{tab:sem_seg_nvs_ours}. \method improves over baseline methods according to most evaluation metrics, underscoring the realism of our re-simulated LiDAR scans.

\subsection{Scene editing}
Beyond LiDAR novel view synthesis by adjusting the sensor configuration (\cf \cref{fig:lidar_nvs}), we also demonstrate the practicality of our compositional neural fields approach with two scene editing applications.

\vspace{-3mm}
\paragraph{Inserting objects from other scenes.}
Our explicit scene decomposition and flexible composition technique enable seamless insertion and removal of neural assets across scenes. As demonstrated in~\cref{fig:vehicle_insertion}, we are able to replace a car from one scene with a truck from another scene, achieving accurate reconstruction of both geometry and intensity. In contrast, UniSim~\cite{yang2023unisim} struggles to preserve high-quality geometry. This highlights the potential of our approach to generate diverse and realistic LiDAR scans of autonomous driving scenarios.

\vspace{-3mm}
\paragraph{Manipulating object trajectories.}
\method also facilitates the manipulation of moving objects' trajectories, by simply adjusting their relative poses to the canonical bounding box. Representative results are shown in ~\cref{fig:traj}. The high realism of our re-simulation is also indicated by the successful detection of inserted virtual objects.

\section{Conclusion and future work}
\vspace{-1mm}
We have presented DyNFL, a compositional neural fields approach for LiDAR re-simulation. Our method surpasses prior art in both static and dynamic scenes, offering powerful scene editing capabilities with the potential to synthesize diverse and high-quality scenes, e.g., to evaluate a perception system trained only on real data in closed-loop mode.

Despite state-of-the-art performance, there are still limitations we aim to address in future work. First, \method has difficulties in synthesizing moving vehicles from unseen view angles. The task is even more challenging than pure shape completion, because the learned prior must include the ability to simulate also intensity, ray drop patterns, \etc in the unseen region. Second, our method currently relies on bounding boxes and trajectories of the moving objects, and its performance may be compromised when bounding boxes are inaccurate. Overcoming this dependency, exploring 4D representations while retaining scene editing flexibility, stands out as a main challenge for future research.
\small
\paragraph{Acknowledgments}
Or Litany is a Taub fellow and is supported by the Azrieli Foundation Early Career Faculty Fellowship.

{\small
\bibliographystyle{ieeenat_fullname}
\bibliography{main}

\begin{thebibliography}{64}
\providecommand{\natexlab}[1]{#1}
\providecommand{\url}[1]{\texttt{#1}}
\expandafter\ifx\csname urlstyle\endcsname\relax
  \providecommand{\doi}[1]{doi: #1}\else
  \providecommand{\doi}{doi: \begingroup \urlstyle{rm}\Url}\fi

\bibitem[Attal et~al.(2023)Attal, Huang, Richardt, Zollhoefer, Kopf, O'Toole, and Kim]{attal2023hyperreel}
Benjamin Attal, Jia-Bin Huang, Christian Richardt, Michael Zollhoefer, Johannes Kopf, Matthew O'Toole, and Changil Kim.
\newblock {HyperReel}: High-fidelity {6-DoF} video with ray-conditioned sampling.
\newblock In \emph{CVPR}, 2023.

\bibitem[Barron et~al.(2021)Barron, Mildenhall, Tancik, Hedman, Martin-Brualla, and Srinivasan]{barron2021mip}
Jonathan~T Barron, Ben Mildenhall, Matthew Tancik, Peter Hedman, Ricardo Martin-Brualla, and Pratul~P Srinivasan.
\newblock Mip-{NeRF}: A multiscale representation for anti-aliasing neural radiance fields.
\newblock In \emph{ICCV}, 2021.

\bibitem[Barron et~al.(2022)Barron, Mildenhall, Verbin, Srinivasan, and Hedman]{barron2022mip}
Jonathan~T Barron, Ben Mildenhall, Dor Verbin, Pratul~P Srinivasan, and Peter Hedman.
\newblock Mip-nerf 360: Unbounded anti-aliased neural radiance fields.
\newblock In \emph{CVPR}, 2022.

\bibitem[Berman et~al.(2018)Berman, Triki, and Blaschko]{berman2018lovasz}
Maxim Berman, Amal~Rannen Triki, and Matthew~B Blaschko.
\newblock The {Lov{\'a}sz}-softmax loss: A tractable surrogate for the optimization of the intersection-over-union measure in neural networks.
\newblock In \emph{CVPR}, 2018.

\bibitem[Caccia et~al.(2019)Caccia, Van~Hoof, Courville, and Pineau]{caccia2019deep}
Lucas Caccia, Herke Van~Hoof, Aaron Courville, and Joelle Pineau.
\newblock Deep generative modeling of lidar data.
\newblock In \emph{IROS}. IEEE, 2019.

\bibitem[Chang et~al.(2023)Chang, Sharma, Kaess, and Lucey]{chang2023neural}
MingFang Chang, Akash Sharma, Michael Kaess, and Simon Lucey.
\newblock Neural radiance field with {LiDAR} maps.
\newblock In \emph{ICCV}, 2023.

\bibitem[Chen et~al.(2022)Chen, Xu, Geiger, Yu, and Su]{chen2022tensorf}
Anpei Chen, Zexiang Xu, Andreas Geiger, Jingyi Yu, and Hao Su.
\newblock {TensorRF}: Tensorial radiance fields.
\newblock In \emph{ECCV}. Springer, 2022.

\bibitem[Deng et~al.(2022)Deng, Liu, Zhu, and Ramanan]{kangle2021dsnerf}
Kangle Deng, Andrew Liu, Jun-Yan Zhu, and Deva Ramanan.
\newblock Depth-supervised {NeRF}: Fewer views and faster training for free.
\newblock In \emph{CVPR}, 2022.

\bibitem[Dosovitskiy et~al.(2017)Dosovitskiy, Ros, Codevilla, Lopez, and Koltun]{dosovitskiy2017carla}
Alexey Dosovitskiy, German Ros, Felipe Codevilla, Antonio Lopez, and Vladlen Koltun.
\newblock Carla: An open urban driving simulator.
\newblock In \emph{Conference on robot learning}. PMLR, 2017.

\bibitem[Fan et~al.(2023)Fan, Wang, Wang, and Zhang]{fan2023fsdv2}
Lue Fan, Feng Wang, Naiyan Wang, and Zhaoxiang Zhang.
\newblock {FSD V2}: Improving fully sparse 3d object detection with virtual voxels.
\newblock \emph{arXiv preprint arXiv:2308.03755}, 2023.

\bibitem[Fang et~al.(2020)Fang, Zhou, Yan, Zhao, Zhang, Ma, Wang, and Yang]{fang2020augmented}
Jin Fang, Dingfu Zhou, Feilong Yan, Tongtong Zhao, Feihu Zhang, Yu Ma, Liang Wang, and Ruigang Yang.
\newblock Augmented lidar simulator for autonomous driving.
\newblock \emph{IEEE Robotics and Automation Letters}, 5\penalty0 (2):\penalty0 1931--1938, 2020.

\bibitem[Fridovich-Keil et~al.(2022)Fridovich-Keil, Yu, Tancik, Chen, Recht, and Kanazawa]{fridovich2022plenoxels}
Sara Fridovich-Keil, Alex Yu, Matthew Tancik, Qinhong Chen, Benjamin Recht, and Angjoo Kanazawa.
\newblock Plenoxels: Radiance fields without neural networks.
\newblock In \emph{CVPR}, 2022.

\bibitem[Gojcic et~al.(2021)Gojcic, Litany, Wieser, Guibas, and Birdal]{gojcic2021weakly}
Zan Gojcic, Or Litany, Andreas Wieser, Leonidas~J Guibas, and Tolga Birdal.
\newblock Weakly supervised learning of rigid 3d scene flow.
\newblock \emph{CVPR 2021}, 2021.

\bibitem[Gropp et~al.(2020)Gropp, Yariv, Haim, Atzmon, and Lipman]{icml2020_2086}
Amos Gropp, Lior Yariv, Niv Haim, Matan Atzmon, and Yaron Lipman.
\newblock Implicit geometric regularization for learning shapes.
\newblock In \emph{Proceedings of Machine Learning and Systems 2020}, pages 3569--3579. 2020.

\bibitem[Guillard et~al.(2022)Guillard, Vemprala, Gupta, Miksik, Vineet, Fua, and Kapoor]{guillard2022learning}
Beno{\^\i}t Guillard, Sai Vemprala, Jayesh~K Gupta, Ondrej Miksik, Vibhav Vineet, Pascal Fua, and Ashish Kapoor.
\newblock Learning to simulate realistic lidars.
\newblock In \emph{IROS}, pages 8173--8180. IEEE, 2022.

\bibitem[Hahner et~al.(2021)Hahner, Sakaridis, Dai, and {Van Gool}]{hahner2021fog}
Martin Hahner, Christos Sakaridis, Dengxin Dai, and Luc {Van Gool}.
\newblock Fog simulation on real {LiDAR} point clouds for 3d object detection in adverse weather.
\newblock In \emph{CVPR}, 2021.

\bibitem[Hahner et~al.(2022)Hahner, Sakaridis, Bijelic, Heide, Yu, Dai, and Van~Gool]{hahner2022lidar}
Martin Hahner, Christos Sakaridis, Mario Bijelic, Felix Heide, Fisher Yu, Dengxin Dai, and Luc Van~Gool.
\newblock {LiDAR} snowfall simulation for robust 3d object detection.
\newblock In \emph{CVPR}, 2022.

\bibitem[Huang et~al.(2022)Huang, Gojcic, Huang, and Andreas~Wieser]{huang2022dynamic}
Shengyu Huang, Zan Gojcic, Jiahui Huang, and Konrad~Schindler Andreas~Wieser.
\newblock Dynamic 3d scene analysis by point cloud accumulation.
\newblock In \emph{ECCV}, 2022.

\bibitem[Huang et~al.(2023)Huang, Gojcic, Wang, Williams, Kasten, Fidler, Schindler, and Litany]{Huang2023nfl}
Shengyu Huang, Zan Gojcic, Zian Wang, Francis Williams, Yoni Kasten, Sanja Fidler, Konrad Schindler, and Or Litany.
\newblock Neural lidar fields for novel view synthesis.
\newblock In \emph{ICCV}, 2023.

\bibitem[Kerbl et~al.(2023)Kerbl, Kopanas, Leimk{\"u}hler, and Drettakis]{kerbl20233d}
Bernhard Kerbl, Georgios Kopanas, Thomas Leimk{\"u}hler, and George Drettakis.
\newblock 3d gaussian splatting for real-time radiance field rendering.
\newblock \emph{ACM Transactions on Graphics}, 42\penalty0 (4):\penalty0 1--14, 2023.

\bibitem[Kingma and Ba(2014)]{kingma2014adam}
Diederik~P Kingma and Jimmy Ba.
\newblock Adam: A method for stochastic optimization.
\newblock \emph{arXiv preprint arXiv:1412.6980}, 2014.

\bibitem[Kundu et~al.(2022)Kundu, Genova, Yin, Fathi, Pantofaru, Guibas, Tagliasacchi, Dellaert, and Funkhouser]{KunduCVPR2022PNF}
Abhijit Kundu, Kyle Genova, Xiaoqi Yin, Alireza Fathi, Caroline Pantofaru, Leonidas Guibas, Andrea Tagliasacchi, Frank Dellaert, and Thomas Funkhouser.
\newblock Panoptic neural fields: A semantic object-aware neural scene representation.
\newblock In \emph{CVPR}, 2022.

\bibitem[Li et~al.(2023{\natexlab{a}})Li, Ren, and Liu]{li2023pcgen}
Chenqi Li, Yuan Ren, and Bingbing Liu.
\newblock Pcgen: Point cloud generator for lidar simulation.
\newblock In \emph{ICRA}, pages 11676--11682. IEEE, 2023{\natexlab{a}}.

\bibitem[Li et~al.(2021)Li, Niklaus, Snavely, and Wang]{li2020neural}
Zhengqi Li, Simon Niklaus, Noah Snavely, and Oliver Wang.
\newblock Neural scene flow fields for space-time view synthesis of dynamic scenes.
\newblock In \emph{CVPR}, 2021.

\bibitem[Li et~al.(2023{\natexlab{b}})Li, M\"uller, Evans, Taylor, Unberath, Liu, and Lin]{li2023neuralangelo}
Zhaoshuo Li, Thomas M\"uller, Alex Evans, Russell~H Taylor, Mathias Unberath, Ming-Yu Liu, and Chen-Hsuan Lin.
\newblock Neuralangelo: High-fidelity neural surface reconstruction.
\newblock In \emph{CVPR}, 2023{\natexlab{b}}.

\bibitem[Liu et~al.(2023)Liu, Gao, Meuleman, Tseng, Saraf, Kim, Chuang, Kopf, and Huang]{liu2023robust}
Yu-Lun Liu, Chen Gao, Andreas Meuleman, Hung-Yu Tseng, Ayush Saraf, Changil Kim, Yung-Yu Chuang, Johannes Kopf, and Jia-Bin Huang.
\newblock Robust dynamic radiance fields.
\newblock In \emph{CVPR}, 2023.

\bibitem[Manivasagam et~al.(2020)Manivasagam, Wang, Wong, Zeng, Sazanovich, Tan, Yang, Ma, and Urtasun]{manivasagam2020lidarsim}
Sivabalan Manivasagam, Shenlong Wang, Kelvin Wong, Wenyuan Zeng, Mikita Sazanovich, Shuhan Tan, Bin Yang, Wei-Chiu Ma, and Raquel Urtasun.
\newblock {LiDARsim}: Realistic {LiDAR} simulation by leveraging the real world.
\newblock In \emph{CVPR}, 2020.

\bibitem[Mildenhall et~al.(2020)Mildenhall, Srinivasan, Tancik, Barron, Ramamoorthi, and Ng]{mildenhall2020nerf}
Ben Mildenhall, Pratul~P Srinivasan, Matthew Tancik, Jonathan~T Barron, Ravi Ramamoorthi, and Ren Ng.
\newblock {NerF}: Representing scenes as neural radiance fields for view synthesis.
\newblock In \emph{ECCV}, 2020.

\bibitem[M\"uller et~al.(2022)M\"uller, Evans, Schied, and Keller]{mueller2022instant}
Thomas M\"uller, Alex Evans, Christoph Schied, and Alexander Keller.
\newblock Instant neural graphics primitives with a multiresolution hash encoding.
\newblock \emph{ACM Trans. Graph.}, 41\penalty0 (4):\penalty0 102:1--102:15, 2022.

\bibitem[Niemeyer et~al.(2020)Niemeyer, Mescheder, Oechsle, and Geiger]{niemeyer2020differentiable}
Michael Niemeyer, Lars Mescheder, Michael Oechsle, and Andreas Geiger.
\newblock Differentiable volumetric rendering: Learning implicit 3d representations without 3d supervision.
\newblock In \emph{CVPR}, 2020.

\bibitem[Oechsle et~al.(2021)Oechsle, Peng, and Geiger]{oechsle2021unisurf}
Michael Oechsle, Songyou Peng, and Andreas Geiger.
\newblock Unisurf: Unifying neural implicit surfaces and radiance fields for multi-view reconstruction.
\newblock In \emph{ICCV}, 2021.

\bibitem[Ost et~al.(2021)Ost, Mannan, Thuerey, Knodt, and Heide]{Ost_2021_CVPR}
Julian Ost, Fahim Mannan, Nils Thuerey, Julian Knodt, and Felix Heide.
\newblock Neural scene graphs for dynamic scenes.
\newblock In \emph{CVPR}, 2021.

\bibitem[Park et~al.(2021{\natexlab{a}})Park, Sinha, Barron, Bouaziz, Goldman, Seitz, and Martin-Brualla]{park2021nerfies}
Keunhong Park, Utkarsh Sinha, Jonathan~T. Barron, Sofien Bouaziz, Dan~B Goldman, Steven~M. Seitz, and Ricardo Martin-Brualla.
\newblock Nerfies: Deformable neural radiance fields.
\newblock In \emph{ICCV}, 2021{\natexlab{a}}.

\bibitem[Park et~al.(2021{\natexlab{b}})Park, Sinha, Hedman, Barron, Bouaziz, Goldman, Martin-Brualla, and Seitz]{park2021hypernerf}
Keunhong Park, Utkarsh Sinha, Peter Hedman, Jonathan~T. Barron, Sofien Bouaziz, Dan~B Goldman, Ricardo Martin-Brualla, and Steven~M. Seitz.
\newblock {HyperNeRF}: A higher-dimensional representation for topologically varying neural radiance fields.
\newblock \emph{ACM Transactions on Graphics}, 40\penalty0 (6), 2021{\natexlab{b}}.

\bibitem[Pumarola et~al.(2021)Pumarola, Corona, Pons-Moll, and Moreno-Noguer]{pumarola2020d}
Albert Pumarola, Enric Corona, Gerard Pons-Moll, and Francesc Moreno-Noguer.
\newblock D-nerf: Neural radiance fields for dynamic scenes.
\newblock In \emph{Proceedings of the IEEE/CVF Conference on Computer Vision and Pattern Recognition}, pages 10318--10327, 2021.

\bibitem[Rematas et~al.(2021)Rematas, Liu, Srinivasan, Barron, Tagliasacchi, Funkhouser, and Ferrari]{rematas2021urban}
Konstantinos Rematas, Andrew Liu, Pratul~P Srinivasan, Jonathan~T Barron, Andrea Tagliasacchi, Thomas Funkhouser, and Vittorio Ferrari.
\newblock Urban radiance fields.
\newblock \emph{arXiv preprint arXiv:2111.14643}, 2021.

\bibitem[Rematas et~al.(2022)Rematas, Liu, Srinivasan, Barron, Tagliasacchi, Funkhouser, and Ferrari]{rematas2022urban}
Konstantinos Rematas, Andrew Liu, Pratul~P Srinivasan, Jonathan~T Barron, Andrea Tagliasacchi, Thomas Funkhouser, and Vittorio Ferrari.
\newblock Urban radiance fields.
\newblock In \emph{CVPR}, 2022.

\bibitem[{Sara Fridovich-Keil and Giacomo Meanti} et~al.(2023){Sara Fridovich-Keil and Giacomo Meanti}, Warburg, Recht, and Kanazawa]{kplanes_2023}
{Sara Fridovich-Keil and Giacomo Meanti}, Frederik~Rahbæk Warburg, Benjamin Recht, and Angjoo Kanazawa.
\newblock K-planes: Explicit radiance fields in space, time, and appearance.
\newblock In \emph{CVPR}, 2023.

\bibitem[Shah et~al.(2018)Shah, Dey, Lovett, and Kapoor]{shah2018airsim}
Shital Shah, Debadeepta Dey, Chris Lovett, and Ashish Kapoor.
\newblock Airsim: High-fidelity visual and physical simulation for autonomous vehicles.
\newblock In \emph{International Conference on Field and Service Robotics}, pages 621--635. Springer, 2018.

\bibitem[Shoemake(1985)]{10.1145/325334.325242}
Ken Shoemake.
\newblock Animating rotation with quaternion curves.
\newblock In \emph{Conference on Computer Graphics and Interactive Techniques}, page 245–254. Association for Computing Machinery, 1985.

\bibitem[Sun et~al.(2022)Sun, Chen, Wang, Li, Averbuch-Elor, Zhou, and Snavely]{sun2022neural}
Jiaming Sun, Xi Chen, Qianqian Wang, Zhengqi Li, Hadar Averbuch-Elor, Xiaowei Zhou, and Noah Snavely.
\newblock Neural 3d reconstruction in the wild.
\newblock In \emph{ACM SIGGRAPH}, 2022.

\bibitem[Sun et~al.(2020)Sun, Kretzschmar, Dotiwalla, Chouard, Patnaik, Tsui, Guo, Zhou, Chai, Caine, et~al.]{sun2020scalability}
Pei Sun, Henrik Kretzschmar, Xerxes Dotiwalla, Aurelien Chouard, Vijaysai Patnaik, Paul Tsui, James Guo, Yin Zhou, Yuning Chai, Benjamin Caine, et~al.
\newblock Scalability in perception for autonomous driving: {Waymo} open dataset.
\newblock In \emph{CVPR}, 2020.

\bibitem[Tagliasacchi and Mildenhall(2022)]{tagliasacchi2022volume}
Andrea Tagliasacchi and Ben Mildenhall.
\newblock Volume rendering digest (for nerf).
\newblock \emph{arXiv preprint arXiv:2209.02417}, 2022.

\bibitem[Tancik et~al.(2023)Tancik, Weber, Ng, Li, Yi, Kerr, Wang, Kristoffersen, Austin, Salahi, Ahuja, McAllister, and Kanazawa]{nerfstudio}
Matthew Tancik, Ethan Weber, Evonne Ng, Ruilong Li, Brent Yi, Justin Kerr, Terrance Wang, Alexander Kristoffersen, Jake Austin, Kamyar Salahi, Abhik Ahuja, David McAllister, and Angjoo Kanazawa.
\newblock Nerfstudio: A modular framework for neural radiance field development.
\newblock In \emph{ACM SIGGRAPH 2023 Conference Proceedings}, 2023.

\bibitem[Tang et~al.(2020)Tang, Liu, Zhao, Lin, Lin, Wang, and Han]{tang2020searching}
Haotian* Tang, Zhijian* Liu, Shengyu Zhao, Yujun Lin, Ji Lin, Hanrui Wang, and Song Han.
\newblock Searching efficient 3d architectures with sparse point-voxel convolution.
\newblock In \emph{ECCV}, 2020.

\bibitem[Tang et~al.(2022)Tang, Liu, Li, Lin, and Han]{tang2022torchsparse}
Haotian Tang, Zhijian Liu, Xiuyu Li, Yujun Lin, and Song Han.
\newblock Torchsparse: Efficient point cloud inference engine.
\newblock \emph{Proceedings of Machine Learning and Systems}, 4:\penalty0 302--315, 2022.

\bibitem[Tao et~al.(2023)Tao, Gao, Wang, Chen, Hao, Liang, Salzmann, and Yu]{tao2023lidar}
Tang Tao, Longfei Gao, Guangrun Wang, Peng Chen, Dayang Hao, Xiaodan Liang, Mathieu Salzmann, and Kaicheng Yu.
\newblock Lidar-nerf: Novel lidar view synthesis via neural radiance fields.
\newblock \emph{arXiv preprint arXiv:2304.10406}, 2023.

\bibitem[Turki et~al.(2023)Turki, Zhang, Ferroni, and Ramanan]{turki2023suds}
Haithem Turki, Jason~Y Zhang, Francesco Ferroni, and Deva Ramanan.
\newblock Suds: Scalable urban dynamic scenes.
\newblock In \emph{Proceedings of the IEEE/CVF Conference on Computer Vision and Pattern Recognition}, pages 12375--12385, 2023.

\bibitem[Verbin et~al.(2022)Verbin, Hedman, Mildenhall, Zickler, Barron, and Srinivasan]{verbin2022ref}
Dor Verbin, Peter Hedman, Ben Mildenhall, Todd Zickler, Jonathan~T Barron, and Pratul~P Srinivasan.
\newblock Ref-nerf: Structured view-dependent appearance for neural radiance fields.
\newblock In \emph{CVPR}, 2022.

\bibitem[Wang et~al.(2021)Wang, Liu, Liu, Theobalt, Komura, and Wang]{wang2021neus}
Peng Wang, Lingjie Liu, Yuan Liu, Christian Theobalt, Taku Komura, and Wenping Wang.
\newblock Neus: Learning neural implicit surfaces by volume rendering for multi-view reconstruction.
\newblock In \emph{NeurIPS}, 2021.

\bibitem[Wang et~al.(2022)Wang, Skorokhodov, and Wonka]{wang2022hf}
Yiqun Wang, Ivan Skorokhodov, and Peter Wonka.
\newblock Hf-neus: Improved surface reconstruction using high-frequency details.
\newblock In \emph{NeurIPS}, 2022.

\bibitem[Wang et~al.(2023{\natexlab{a}})Wang, Han, Habermann, Daniilidis, Theobalt, and Liu]{wang2023neus2}
Yiming Wang, Qin Han, Marc Habermann, Kostas Daniilidis, Christian Theobalt, and Lingjie Liu.
\newblock Neus2: Fast learning of neural implicit surfaces for multi-view reconstruction.
\newblock In \emph{ICCV}, 2023{\natexlab{a}}.

\bibitem[Wang et~al.(2023{\natexlab{b}})Wang, Shen, Gao, Huang, Munkberg, Hasselgren, Gojcic, Chen, and Fidler]{wang2023neural}
Zian Wang, Tianchang Shen, Jun Gao, Shengyu Huang, Jacob Munkberg, Jon Hasselgren, Zan Gojcic, Wenzheng Chen, and Sanja Fidler.
\newblock Neural fields meet explicit geometric representations for inverse rendering of urban scenes.
\newblock In \emph{CVPR}, 2023{\natexlab{b}}.

\bibitem[Wu et~al.(2022{\natexlab{a}})Wu, Wang, Pan, Xu, Theobalt, Liu, and Lin]{wu2022voxurf}
Tong Wu, Jiaqi Wang, Xingang Pan, Xudong Xu, Christian Theobalt, Ziwei Liu, and Dahua Lin.
\newblock Voxurf: Voxel-based efficient and accurate neural surface reconstruction.
\newblock \emph{arXiv preprint arXiv:2208.12697}, 2022{\natexlab{a}}.

\bibitem[Wu et~al.(2022{\natexlab{b}})Wu, Zhong, Tagliasacchi, Cole, and Oztireli]{wu2022d}
Tianhao Wu, Fangcheng Zhong, Andrea Tagliasacchi, Forrester Cole, and Cengiz Oztireli.
\newblock D\^{} 2nerf: Self-supervised decoupling of dynamic and static objects from a monocular video.
\newblock \emph{Advances in Neural Information Processing Systems}, 35:\penalty0 32653--32666, 2022{\natexlab{b}}.

\bibitem[Xie et~al.(2022)Xie, Takikawa, Saito, Litany, Yan, Khan, Tombari, Tompkin, Sitzmann, and Sridhar]{xie2022neural}
Yiheng Xie, Towaki Takikawa, Shunsuke Saito, Or Litany, Shiqin Yan, Numair Khan, Federico Tombari, James Tompkin, Vincent Sitzmann, and Srinath Sridhar.
\newblock Neural fields in visual computing and beyond.
\newblock In \emph{Computer Graphics Forum}, pages 641--676, 2022.

\bibitem[Yang et~al.(2023{\natexlab{a}})Yang, Ivanovic, Litany, Weng, Kim, Li, Che, Xu, Fidler, Pavone, and Wang]{yang2023emernerf}
Jiawei Yang, Boris Ivanovic, Or Litany, Xinshuo Weng, Seung~Wook Kim, Boyi Li, Tong Che, Danfei Xu, Sanja Fidler, Marco Pavone, and Yue Wang.
\newblock Emernerf: Emergent spatial-temporal scene decomposition via self-supervision.
\newblock \emph{arXiv preprint arXiv:2311.02077}, 2023{\natexlab{a}}.

\bibitem[Yang et~al.(2023{\natexlab{b}})Yang, Chen, Wang, Manivasagam, Ma, Yang, and Urtasun]{yang2023unisim}
Ze Yang, Yun Chen, Jingkang Wang, Sivabalan Manivasagam, Wei-Chiu Ma, Anqi~Joyce Yang, and Raquel Urtasun.
\newblock Unisim: A neural closed-loop sensor simulator.
\newblock In \emph{CVPR}, 2023{\natexlab{b}}.

\bibitem[Yariv et~al.(2021)Yariv, Gu, Kasten, and Lipman]{yariv2021volume}
Lior Yariv, Jiatao Gu, Yoni Kasten, and Yaron Lipman.
\newblock Volume rendering of neural implicit surfaces.
\newblock In \emph{NeurIPS}, 2021.

\bibitem[Yu et~al.(2022)Yu, Peng, Niemeyer, Sattler, and Geiger]{yu2022monosdf}
Zehao Yu, Songyou Peng, Michael Niemeyer, Torsten Sattler, and Andreas Geiger.
\newblock {MonoSDF}: Exploring monocular geometric cues for neural implicit surface reconstruction.
\newblock In \emph{NeurIPS}, 2022.

\bibitem[Yuan et~al.(2021)Yuan, Lv, Schmidt, and Lovegrove]{yuan2021star}
Wentao Yuan, Zhaoyang Lv, Tanner Schmidt, and Steven Lovegrove.
\newblock Star: Self-supervised tracking and reconstruction of rigid objects in motion with neural rendering.
\newblock In \emph{CVPR}, 2021.

\bibitem[Zhang et~al.(2023)Zhang, Zhang, Kuang, and Zhang]{zhang2023nerf}
Junge Zhang, Feihu Zhang, Shaochen Kuang, and Li Zhang.
\newblock Nerf-lidar: Generating realistic lidar point clouds with neural radiance fields.
\newblock \emph{arXiv preprint arXiv:2304.14811}, 2023.

\bibitem[Zuo et~al.(2023)Zuo, Yang, Merrill, Xu, and Leutenegger]{zuo2023incremental}
Xingxing Zuo, Nan Yang, Nathaniel Merrill, Binbin Xu, and Stefan Leutenegger.
\newblock Incremental dense reconstruction from monocular video with guided sparse feature volume fusion.
\newblock \emph{IEEE Robotics and Automation Letters}, 2023.

\bibitem[Zyrianov et~al.(2022)Zyrianov, Zhu, and Wang]{zyrianov2022learning}
Vlas Zyrianov, Xiyue Zhu, and Shenlong Wang.
\newblock Learning to generate realistic lidar point clouds.
\newblock In \emph{ECCV}. Springer, 2022.

\end{thebibliography}
}
\clearpage
\setcounter{section}{0}
\renewcommand\thesection{\Alph{section}}
\newcommand{\manuallabel}[2]{\def\@currentlabel{#2}\label{#1}}
\makeatother
\newcommand{\refpaper}[1]{{\hypersetup{linkcolor={red}}\ref{#1}}}
\allowdisplaybreaks
In this supplementary material, we first provide additional information about the datasets for our evaluations and implementation details of our proposed method in~\cref{sec:sup_dataset}. Next, we present more qualitative and quantitative results in~\cref{sec:sup_moreresults}. Please also check the supplemental video for more results showcasing our performance. Finally, we provide the complete derivations of the SDF-based volume rendering for active sensor in~\cref{sec:sup_sdf_vol_render}. 

\section{Datasets and implementation details}\label{sec:sup_dataset}
\subsection{Datasets}
\paragraph{\textit{Waymo Dynamic.}} For the \textit{Waymo Dynamic} dataset, we take them from 4 scenes of \textit{Waymo Open Dataset}~\cite{sun2020scalability}. There are multiple moving vehicles inside each scene. 50 consecutive frames are taken from each scene for our evaluation. The vehicles are deemed as \textit{dynamic} if the speed is $>1\,$m/s. in any of the 50 frames. The corresponding scene IDs on \textit{Waymo Open Dataset} for our selected scenes are shown as follows:
\begin{table}[!h]
    \setlength{\tabcolsep}{4pt}
    \renewcommand{\arraystretch}{1.2}
	\centering
	\resizebox{0.8\columnwidth}{!}{
    \begin{tabular}{l|c}
    \toprule
    & Scene ID \\
    \midrule
    Scene 1 & 1083056852838271990\_4080\_000\_4100\_000 \\
    Scene 2 & 13271285919570645382\_5320\_000\_5340\_000 \\
    Scene 3 & 10072140764565668044\_4060\_000\_4080\_000 \\
    Scene 4 & 10500357041547037089\_1474\_800\_1494\_800 \\
    \bottomrule
    \end{tabular}
    }
\end{table}

\paragraph{\textit{Waymo Dynamic NVS.}} For the \textit{Waymo Dynamic NVS} dataset, we use the same 4 scenes as chosen in \textit{Waymo Dynamic}. We change the evaluation paradigm similar to \textit{Waymo NVS} ~\cite{Huang2023nfl} such that we first train the model on all 50 consecutive LiDAR frames then we synthesize 50 novel LiDAR frames with a sensor shift of 2 meters. We then train a new model on the new 50 synthetic LiDAR scans and evaluate against the original 50 LiDAR scans.

\subsection{Implementation details}
\paragraph{DynNFL.} 
Our model is implemented based on nerfstudio\cite{nerfstudio}. For the static neural field, we sample $N_s=512$ points in total, with $N_u=256$ uniformly sampled points and $N_i=256$ weighted sampled points with 8 upsample steps. In each upsample step, 32 points are sampled based on the weight distribution of the previously sampled points. For each dynamic neural field, we sample $N_s=128$ points in total, with $N_u=64$ uniformly sampled points and $N_i=64$ weighted sampled points with 4 upsample steps. During training, we minimize the loss function using the Adam~\cite{kingma2014adam} optimiser, with an initial learning rate of 0.005. It linearly decays to 0.0005 towards the end of training. For the loss weights, we use $w_{\zeta}=3, w_{e}=50, w_{\text{drop}}=0.15, w_{s}=1$, and  $w_{\text{eik}}=0.3$. The batch size is 4096 and we train the model for 60000 iterations on a single RTX3090 GPU with float32 precision.

\paragraph{LiDARsim.} We re-implement the LiDARsim~\cite{manivasagam2020lidarsim} as one of our baselines. 
First, we estimated point-wise normal vectors by considering all points within a 20 cm radius ball within the training set. Following this, we applied voxel down-sampling~\cite{tang2022torchsparse}, employing a 4 cm voxel size to reconstruct individual disk surfels at each point. The surfel orientation is defined based on the estimated normal vector. During inference, we apply the ray-surfel intersections test to determine the intersection points, thus the range and intensity values. We select a fixed surfel radius of 6 cm for the \textit{Waymo} dataset and 12 cm for the \textit{Town} dataset.
To handle dynamic vehicles, we follow LiDARsim~\cite{manivasagam2020lidarsim} by aggregating the LiDAR points for each vehicle from all the training frames and representing them in the \textit{canonical} frame of each vehicle. During inference, we transform all the aggregated vehicle points from their \textit{canonical} frames to the world frame and run ray-surfel intersection.

\paragraph{UniSim.} 
We re-implement UniSim's~\cite{yang2023unisim} rendering process for LiDAR measurements by replacing our ray-drop test-based neural fields composition method with its joint rendering method. For every ray $\mathbf{r} (\mathbf{o},\mathbf{d})$, we begin by conducting an intersection test with all dynamic bounding boxes in the scene to identify the near and far limits. We then uniformly sample 512 points along each ray, assigning each point to either a dynamic neural field, if it falls within a dynamic bounding box, or to the static neural field otherwise. After sampling, we query the SDF and intensity values from the relevant neural fields. Finally, using the SDF-based volume rendering formula in Eq.~\ref{eq:depth_render} for active sensors, we calculate the weights and perform the rendering. Note that we use the same neural field architecture as in our method.
\section{Additional results}\label{sec:sup_moreresults}
\subsection{Waymo Dynamic NVS evaluation} To demonstrate the robustness of our method, we extend the evaluation paradigm to not only focus on interpolation performance. We incorporate Waymo NVS evaluation introduced in \cref{sec:sup_dataset} to focus on close-loop novel view synthesis performance. As illustrated in \cref{tab:waymodynamicnvs}, our method outperforms LiDARsim and Unisim in all aspects.

\subsection{Future frame generation}
\begin{figure}[t]
    \centering
        \includegraphics[width=1\linewidth]{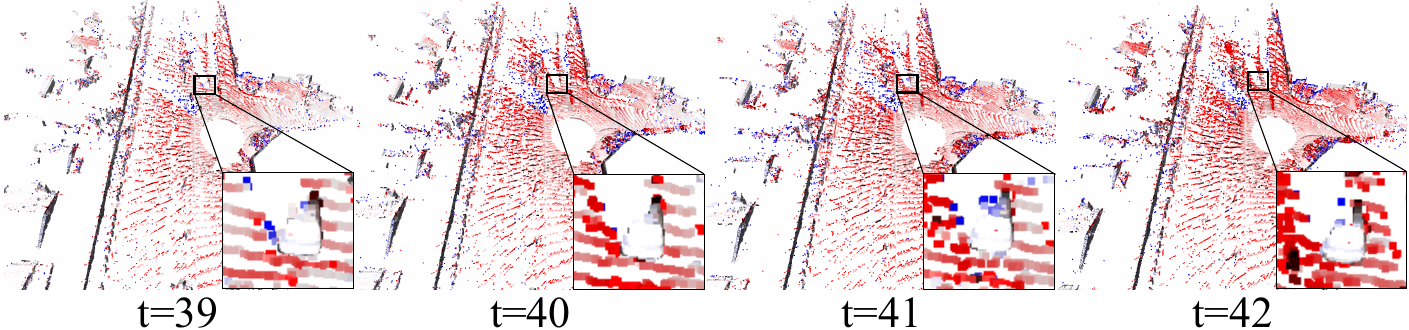}
        \caption{Qualitative results of LiDAR future frame simulation.
        }
    \label{fig:future_frames}
\end{figure}

We trained DyNFL using the initial 40 frames and assessed its performance against the last 10 frames. The results are presented on the \textit{Waymo Dynamic} dataset in \cref{tab:future_frame_quant} and \cref{fig:future_frames}. Unsurprisingly, the performance is comparatively inferior to the original setting (\cf Tab.~1), as it requires extrapolation beyond the observed environment, and thus again a (possibly learned) scene prior. Nevertheless DyNFL continues to outperform LiDARsim. The degradation on dynamic vehicles is marginal, attributable to our precise pose interpolation and high-quality asset reconstruction. We will incorporate these findings in the final version.

\begin{table}[t]
    \setlength{\tabcolsep}{4pt}
    \renewcommand{\arraystretch}{1.2}
	\centering
	\resizebox{\columnwidth}{!}{
    \begin{tabular}{l|ccccc}
    \toprule
    Method  & MAE $\downarrow$ &  MedAE $\downarrow$ & CD $\downarrow$ & MedAE Dyn $\downarrow$ & Intensity RMSE $\downarrow$\\
    \midrule
    LiDARsim~\cite{manivasagam2020lidarsim} & 448.4 & 55.1 & 77.0 &  38.7  & 0.13\\
    Unisim~\cite{yang2023unisim} & 115.1 & 9.7 & 33.5 & 24.3 & 0.19\\
    Ours~ & \textbf{72.9} & \textbf{3.8} & \textbf{22.9} &\textbf{14.0} & \textbf{0.07}\\
    \bottomrule
    \end{tabular}
    }
	\caption{Evaluation of LiDAR NVS on \textit{Waymo Dynamic NVS}.}
	\label{tab:waymodynamicnvs}
\end{table}

\begin{table}[t]
    \setlength{\tabcolsep}{4pt}
    \renewcommand{\arraystretch}{1.2}
	\centering
	\resizebox{0.8\columnwidth}{!}{
    \begin{tabular}{l|cccc}
    \toprule
    Method  & MAE $\downarrow$ &  MedAE $\downarrow$ & CD $\downarrow$ & MedAE Dyn $\downarrow$ \\
    \midrule
    LiDARsim & 333.3 & 25.3 & 67.8&  13.0  \\
    Ours~ & \textbf{81.8} & \textbf{8.6} & \textbf{26.4} &\textbf{9.3} \\
    \bottomrule
    \end{tabular}
    }
	\caption{Results of future frame simulation.}
	\label{tab:future_frame_quant}
\end{table}

\subsection{Runtime analysis}
DyNFL training takes $\approx$7 hours on average on a single RTX 3090 GPU with fp16 precision and 16 hours with fp32 prevision, inference takes 2.2 seconds per LiDAR scan using fp16 precision and 7 seconds using fp32 precision. The envisioned offline use for counterfactual re-simulation prioritizes realism over efficiency. Runtime can potentially be improved for high-throughput applications by reducing rendering complexity.

\subsection{More qualitative results}
\begin{figure*}[t!]
  \centering
   \includegraphics[width=1\textwidth]{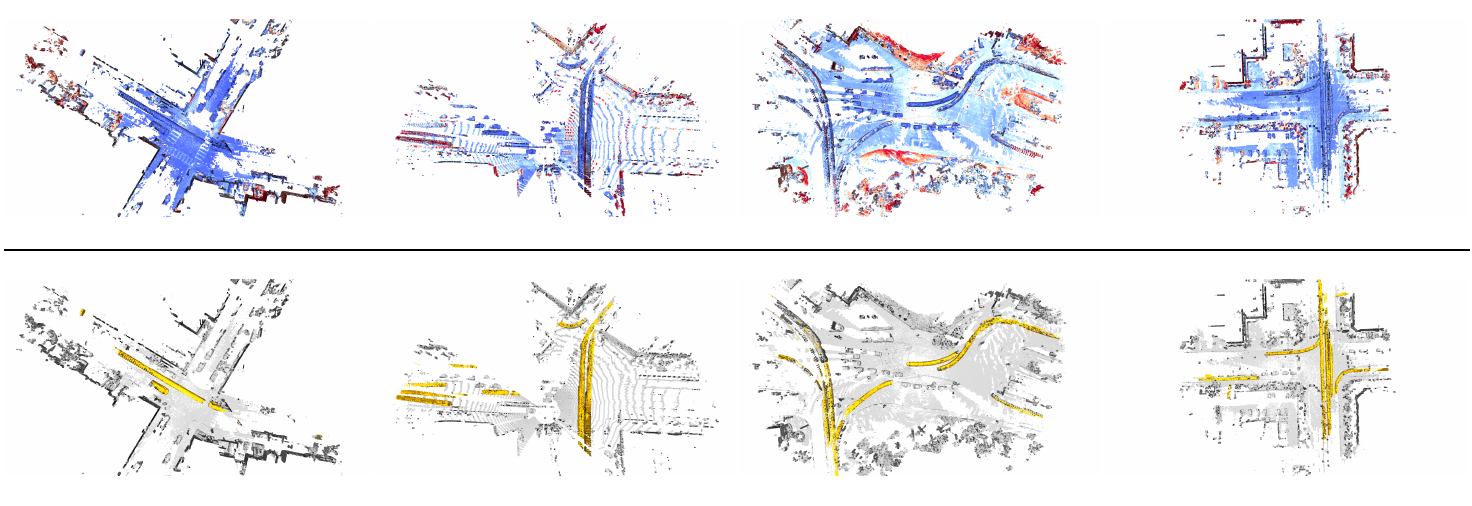}
   \caption{Visualization of 4 selected scenes from \textit{Waymo Dynamic} dataset. For each scene, we aggregate 50 frames. In the first row, points are color-coded by the intensity values(0 ~\bwr~ 0.25). In the second row, dynamic vehicles are painted as \textcolor{yellow}{yellow}.}
   \label{fig:4_scenes_supp}
\end{figure*}

\begin{figure*}[t!]
  \centering
   \includegraphics[width=1\textwidth]{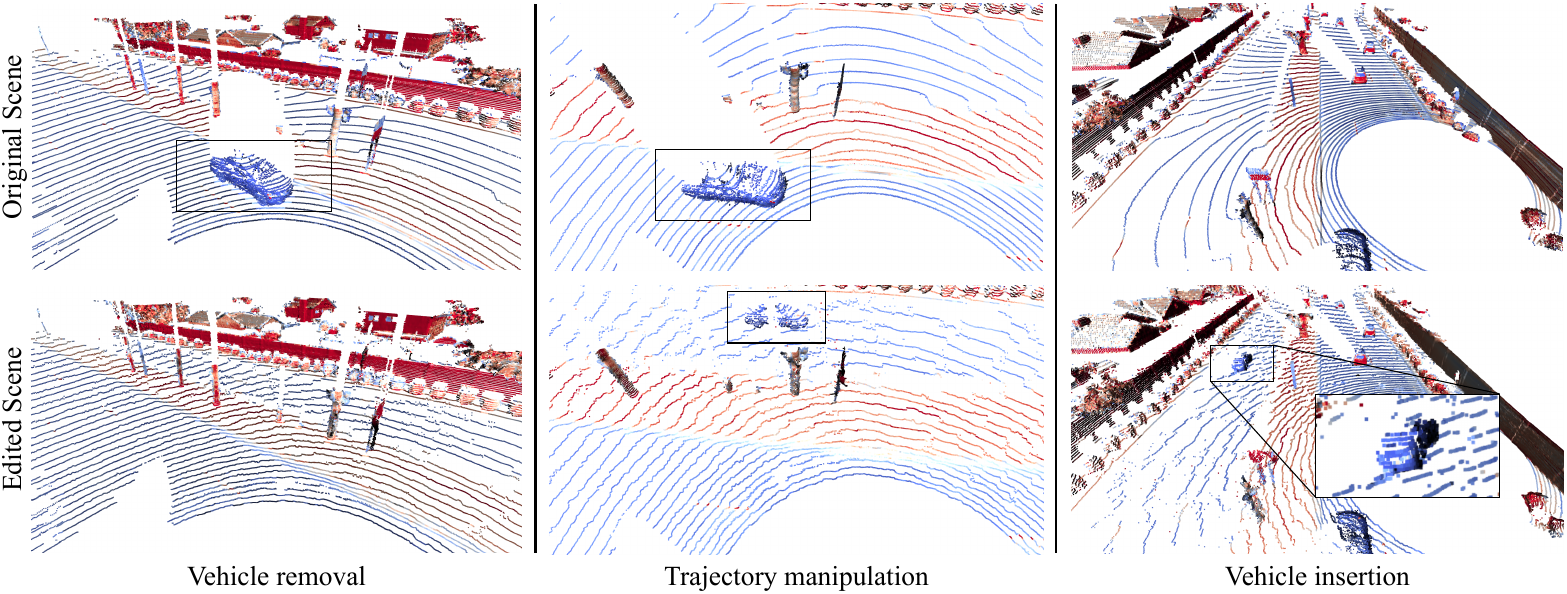}
   \caption{Visualization of scene editing capabilities. We showcase 3 kinds of scene editing capabilities including vehicle removal(left), trajectory manipulation(middle) and vehicle insertion(right). The first row represents the original scenes, the second row demonstrates the scenes after editing. All points are color-coded by the intensity values(0 ~\bwr~ 0.25).}
   \label{fig:scene_editing_supp}
\end{figure*}

In this section, we provide more qualitative results. In \cref{fig:4_scenes_supp}, we showcase the 4 scenes from \textit{Waymo dynamic} dataset. We show additional scene editing results in~\cref{fig:scene_editing_supp}. Please check the supplementary videos for more qualitative results. 

\section{SDF-based LiDAR volume rendering}\label{sec:sup_sdf_vol_render}
In this section, we start by introducing the preliminary of NeRF~\cite{mildenhall2020nerf} following terminology as described in~\cite{tagliasacchi2022volume}. Then we provide the full derivation of the SDF-based volume rendering for active sensor. 

\subsection{Preliminary}\label{sec:supp_pre}
\paragraph{Density.}
For a ray emitted from the origin $\origin \in \real^3$ towards direction $\dir \in \real^3$, the \textit{density} $\density_\zeta$ at range $\zeta$ indicates the likelihood of light interacting with particles at that point $\ray_\zeta = \origin + \zeta \dir$. This interaction can include absorption or scattering of light. In passive sensing, density $\density$ is a critical factor in determining how much light from the scene's illumination is likely to reach the sensor after passing through the medium.
\paragraph{Transmittance}quantifies the likelihood of light traveling through a given portion of the medium without being scattered or absorbed. Density is closely tied to the transmittance function $\trans(\zeta)$, which indicates the probability of a ray traveling over the interval $[0, \zeta)$ without hitting any particles. Then the probability $\trans(\zeta {+} d\zeta)$ of \emph{not} hitting a particle when taking a differential step $d\zeta$ is equal to $\trans(\zeta)$, the likelihood of the ray reaching $\zeta$, times $(1 - d\zeta \cdot \density(\zeta))$, the probability of not hitting anything during the step:
\begin{align}
\trans(\zeta+d\zeta) =& \trans(\zeta) \cdot (1 - d\zeta \cdot \density(\zeta))
\\
\frac{\trans(\zeta+d\zeta) - \trans(\zeta)}{d\zeta} \equiv& \trans'(\zeta) = -\trans(\zeta) \cdot \sigma(\zeta) \;.
\label{eq:derivative}
\end{align}
We solve the differential equation as follows:
\begin{align}
\trans'(\zeta) &= -\trans(\zeta) \cdot \density(\zeta) \\
\frac{\trans'(\zeta)}{\trans(\zeta)} &= -\density(\zeta) \\
\int_a^b \frac{\trans'(\zeta)}{\trans(\zeta)} \; d\zeta &= -\int_a^b \density(\zeta) \; d\zeta \\
\left. \log \trans(\zeta) \right|_a^b &= -\int_a^b \density(\zeta) \; d\zeta \\
\trans(a \rightarrow b) \equiv \frac{\trans(b)}{\trans(a)} &= \exponential{-\int_a^b \density(\zeta) \; d\zeta}   \;.
\end{align}
Hence, for a ray segment between $\zeta_0$ and $\zeta$, transmittance is given by:

\begin{equation}
\trans_{\zeta_0 \rightarrow \zeta} \equiv \frac{\trans_{\zeta}}{\trans_{\zeta_0}} = exp({-\int_{\zeta_0}^\zeta \density_t dt})\;,
\label{eq:trans_ab}
\end{equation}
which leads to following factorization of the transmittance:
\begin{equation}
\trans_{\zeta} = \trans_{0 \rightarrow \zeta_0} \cdot \trans_{\zeta_0 \rightarrow \zeta}\;.
\label{eq:factor}
\end{equation}

\paragraph{Opacity}is the complement of transmittance and represents the fraction of light that is either absorbed or scattered in the medium. In a homogeneous medium with constant density $\density$  the opacity for a segment $[\zeta_j, \zeta_{j+1}]$ of length $\Delta \zeta$ is given by $\opacity_{\zeta_j} = 1 - exp(-\density \cdot \Delta \zeta)$
\subsection{SDF-based volume rendering for active sensor}\label{sec:sdf_active}
NeuS\cite{wang2021neus} derives the opaque density based on the SDF which is:
\begin{equation}
\begin{split}
\density_{\zeta_i} =&  \max\left(\frac{-\frac{d\Phi_s}{d\zeta_i}(f(\zeta_i))}{\Phi_s(f(\zeta_i))},0\right)\\
                  =& \max\left(\frac{-(\nabla f(\zeta_i)\cdot \mathbf{v})\phi_s(f(\zeta_i))}{\Phi_s(f(\zeta_i))}, 0\right) \;,
\end{split}
\label{eq:sigmoid_density_supp}
\end{equation}
where $\Phi_s$ represents the Sigmoid function, $f$ is the SDF function that maps a range $\zeta$ to the SDF value of the point position $\origin + \dir * \zeta$. Note that the integral term is computed by
\begin{equation}
\int \frac{-(\nabla f(\zeta)\cdot \mathbf{v})\phi_s(f(\zeta))}{\Phi_s(f(\zeta))}d\zeta = -\ln(\Phi_s(f(\zeta))) + C \;.
\label{eq:intergration_density}
\end{equation}

We extend the density-based volume rendering for active sensor to SDF-based. Starting from the passive SDF-based volume rendering \cite{wang2021neus}, We substitute the density $\tilde{\density}$ with opaque density in \ref{eq:sigmoid_density_supp}
and evaluate the radiant power integrated from ray segment [a,b] with constant reflectivity $\reflectivity_a$.

Consider the case where $-(\nabla f(\zeta)\cdot \mathbf{v})>0$ within the ray segment $[a,b]$. We have
\begin{align}
P(a \rightarrow b)
&= \int_a^b \trans^2(a\rightarrow t) \cdot \tilde{\density}_t \cdot \reflectivity(t)  \; dt
\\
&= \reflectivity_a \int_a^b \trans^2(a\rightarrow t) \cdot \tilde{\density}_t \; dt
\\
&= \reflectivity_a \int_a^b \exponential{-\int_a^t 2\tilde{\density}(u) \; du} \cdot \tilde{\density}_t \; dt
\\
&= \reflectivity_a \int_a^b \exponential{-2\int_a^t \tilde{\density}(u) \; du} \cdot \tilde{\density}_t \; dt
\\
&= \reflectivity_a \int_a^b \exponential{\left. 2\ln(\Phi_s(f(u)))\right|_a^t} \cdot \tilde{\density}_t \; dt\;.
\end{align}
Let  $\Omega_x = \Phi_s(f(x))$, then
\begin{align}
P(a \rightarrow b)
&= \reflectivity_a \int_a^b \exponential{2\ln(\Omega_t) - 2\ln(\Omega_a)} \cdot \tilde{\density}_t \; dt 
\\
&= \reflectivity_a \int_a^b \frac{{\Omega_t}^2}{{\Omega_a}^2} \cdot \tilde{\density}_t \; dt
\\
&= \frac{\reflectivity_a}{{\Omega_a}^2} \int_a^b {\Omega_t}^2 \cdot \tilde{\density}_t \; dt 
\\
&= \frac{\reflectivity_a}{{\Omega_a}^2} \int_a^b -\frac{d\Phi_s}{dt}(f(t)) \cdot \Phi_s(f(t)) \; dt 
\\
&= \frac{\reflectivity_a}{{\Omega_a}^2} ( \left. -\frac{1}{2}{\Phi_s(f(t))}^2 \right|_a^b) \\
&= \frac{\reflectivity_a}{{\Omega_a}^2} (\frac{1}{2}{\Phi_s(f(a))}^2 -\frac{1}{2}{\Phi_s(f(b))}^2 )\\
&= \frac{{\Phi_s(f(a))}^2 -{\Phi_s(f(b))}^2}{{2\Phi_s(f(a))}^2} \cdot \reflectivity_a \;.
\label{eq:homogeneous}
\end{align}

Consider the case where $-(\nabla f(\zeta)\cdot \mathbf{v})<0$ within the ray segment $[a,b]$. Then
\begin{align}
P(a \rightarrow b)
&= \int_a^b \trans^2(a\rightarrow t) \cdot \tilde{\density}_t \cdot \reflectivity(t)  \; dt
\\
&= \int_a^b \trans^2(a\rightarrow t) \cdot 0 \cdot \reflectivity(t)  \; dt
\\
&= 0 \;.
\end{align}
Hence we conclude 
\begin{align}
P(a \rightarrow b)
&= \max\left(\frac{{\Phi_s(f(a))}^2 -{\Phi_s(f(b))}^2}{{2\Phi_s(f(a))}^2},0\right) \cdot \reflectivity_a \;.
\end{align}

\paragraph{Volume rendering of piecewise constant data.}
Combining the above, we can evaluate the volume rendering integral through a medium with piecewise constant reflectivity:
\begin{align}
P(\zeta_{N+1}) &= \sum_{n=1}^N \int_{\zeta_n}^{\zeta_{n+1}} \trans^2(\zeta) \cdot \tilde{\density}_{\zeta} \cdot \reflectivity_{\zeta_n} \; d\zeta
\\
&= \sum_{n=1}^N \int_{\zeta_n}^{\zeta_{n+1}} \trans^2_{\zeta_n} \cdot \trans^2(\zeta_n \shortto \zeta) \cdot \tilde{\density}_{\zeta} \cdot \reflectivity_{\zeta_n} \; d\zeta 
\\
&= \sum_{n=1}^N \trans^2_{\zeta_n}  \int_{\zeta_n}^{\zeta_{n+1}} \trans^2(\zeta_n \rightarrow \zeta) \cdot \tilde{\density}_{\zeta} \cdot \reflectivity_{\zeta_n} \; d\zeta \\
&=\sum_{n=1}^N \trans^2_{\zeta_n} P(\zeta_n \rightarrow \zeta_{n+1})
\\
&= \sum_{n=1}^N \trans^2_{\zeta_n} \cdot \tilde{\weight}_{\zeta_n} \cdot \reflectivity_{\zeta_n},
\end{align}

where 
\begin{align}
\tilde{\weight}_{\zeta_n} \equiv \max\left(\frac{{\Phi_s(f(\zeta_n)}^2 -{\Phi_s(f(\zeta_{n+1}))}^2}{{2\Phi_s(f(\zeta_n))}^2},0\right) \;.
\end{align}

The discrete accumulated transmittance $\trans$ can be calculated as follows:

Consider the case where $-(\nabla f(\zeta)\cdot \mathbf{v}) > 0$ in $[\zeta_n, \zeta_{n+1}]$: 
\begin{align}
\trans_{\zeta_n} 
&=\prod_{i=1}^{n-1}(\exp(-\int_{\zeta_n}^{\zeta_{n+1}}\tilde{\density}_\zeta \; d\zeta) \\
&= \prod_{i=1}^{n-1}(\frac{\Phi_s(f(\zeta_{n+1}))}{\Phi_s(f(\zeta_n))})\\
\trans^2_{\zeta_n}
&= \prod_{i=1}^{n-1}(\frac{{\Phi_s(f(\zeta_{n+1}))}^2}{{\Phi_s(f(\zeta_n))}^2})\\
&= \prod_{i=1}^{n-1}(1-2\tilde{\weight}_{\zeta_n}) \;.
\label{eq:dicrete_transmittance}
\end{align}

Consider the case where $-(\nabla f(\zeta)\cdot \mathbf{v}) < 0$ in $[\zeta_n, \zeta_{n+1}]$: 
\begin{align}
\trans_{\zeta_n} 
&=\prod_{i=1}^{n-1}(\exp(-\int_{\zeta_n}^{\zeta_{n+1}}\tilde{\density}_\zeta \; d\zeta) = \prod_{i=1}^{n-1}(1)
\\
\trans^2_{\zeta_n} &= \prod_{i=1}^{n-1}(1^2) = \prod_{i=1}^{n-1}(1-2\tilde{\weight}_{\zeta_n}) \;.
\end{align}
In conclusion, the radiant power can be reformulated as:

\begin{align}
P(\zeta_{N+1}) = \sum_{n=1}^N \trans^2_{\zeta_n} \cdot \tilde{\weight}_{\zeta_n} \cdot \reflectivity_{\zeta_n} \;,
\label{eq:final_radiant2}
\end{align}
where $\trans^2_{\zeta_n} = \prod_{i=1}^{n-1}(1-2\tilde{\weight}_{\zeta_i})\;$.

\paragraph{Depth volume rendering of piecewise constant data}

Note that $\tilde{\weight}_{\zeta_n} \in [0, 0.5], \trans^2_{\zeta_n} \in [0,1], \sum_{n=1}^N \trans^2_{\zeta_n} \cdot \tilde{\weight}_{\zeta_n} = 0.5$, for depth volumetric rendering, we have 
\begin{align}
    \zeta = \sum_{n=1}^N 2 \cdot \trans^2_{\zeta_n} \cdot \tilde{\weight}_{\zeta_n} \cdot \zeta_n
    =\sum_{n=1}^N w_n \cdot \zeta_n \;,
    \label{eq:depth_render}
\end{align}
where $w_n = 2\tilde{\weight}_{\zeta_n} \cdot \prod_{i=1}^{n-1}(1-2\tilde{\weight}_{\zeta_i})\;$.

\end{document}